\documentclass{article} %
\usepackage{iclr2023_conference,times}

\usepackage{amsmath,amsfonts,bm}

\def\figref#1{Figure~\ref{#1}}
\def\Figref#1{Figure~\ref{#1}}

\def\secref#1{Section~\ref{#1}}
\def\Secref#1{Section~\ref{#1}}

\def\eqref#1{Eqn.~\ref{#1}}

\def\1{\bm{1}}

\DeclareMathAlphabet{\mathsfit}{\encodingdefault}{\sfdefault}{m}{sl}
\SetMathAlphabet{\mathsfit}{bold}{\encodingdefault}{\sfdefault}{bx}{n}

\usepackage{xcolor}
\definecolor{black}{rgb}{0,0,0}
\usepackage[colorlinks, linkcolor=black, urlcolor=black, citecolor=black]{hyperref}
\usepackage{titlesec}
\titlespacing*{\paragraph}{0pt}{.5ex}{1ex}
\usepackage{url}
\usepackage{cleveref}

\usepackage{tikz}
\usetikzlibrary{shapes}
\usetikzlibrary{graphs,graphs.standard,quotes}

\usepackage{ifthen}
\usepackage{times}
\usepackage{latexsym}
\usepackage{graphicx}
\usepackage{amsmath}
\usepackage{amsfonts}
\usepackage{booktabs}
\usepackage{subcaption}
\usepackage{stringstrings}
\usepackage{siunitx}
\usepackage{algorithm}
\usepackage{algorithmic}
\usepackage{hyperref}
\usepackage{url}
\usepackage{pifont}
\usepackage{booktabs}
\usepackage{marvosym}
\usepackage{tikz}
\usetikzlibrary{positioning}
\usepackage{colortbl}
\usepackage{caption}
\usepackage{multirow}
\usepackage{subcaption}
\usepackage{inconsolata}
\usepackage{stmaryrd}

\newcommand{\CPMIN}{\ensuremath{\text{\OurMethodAbbr{}}_{\text{\InputLevelAbbr}}}}
\newcommand{\CPMHI}{\ensuremath{\text{\OurMethodAbbr{}}_{\text{\NonInputLevelAbbr}}}}

\newcommand{\VarBigU}{U}
\newcommand{\VarBigV}{V}
\newcommand{\VarBigH}{H}
\newcommand{\VarBigX}{X}
\newcommand{\VarU}{u}
\newcommand{\VarV}{v}
\newcommand{\VarToken}{t}
\newcommand{\VarUEdit}{u'}
\newcommand{\VarVEdit}{v'}
\newcommand{\VarInput}{x}
\newcommand{\VarInputAppr}{\ac{x}}
\newcommand{\VarSourceInputAbstract}{x}

\newcommand{\VarSourceInput}{s}
\newcommand{\VarConceptAbstract}{C}
\newcommand{\VarConceptAbstractSingle}{\VarConceptAbstract_{i}}
\newcommand{\VarConcept}{c}
\newcommand{\VarConceptSingle}{\VarConcept_{i}}
\newcommand{\VarConceptEdit}{c'}
\newcommand{\IntInvSign}{\leftarrow}
\newcommand{\Pairs}[2]{({#1}, {#2})}
\newcommand{\AssignFunc}[2]{{#1} = {#2}}
\newcommand{\EditFunc}[2]{{#1} \IntInvSign {#2}}

\newcommand{\UAssign}{\AssignFunc{\VarBigU}{\VarU}}
\newcommand{\VAssign}{\AssignFunc{\VarBigV}{\VarV}}
\newcommand{\UVWorld}{\VarU, \VarV}
\newcommand{\UVWorldPairs}{\Pairs{\VarU}{\VarV}}
\newcommand{\UVEditWorld}{\VarUEdit, \VarVEdit}

\newcommand{\ConceptAssign}{\AssignFunc{\VarConceptAbstractSingle}{\VarConceptEdit}}
\newcommand{\ConceptEdit}{\EditFunc{\VarConceptAbstractSingle}{\VarConceptEdit}}
\newcommand{\Factual}{\VarInput_{\UVWorld}}
\newcommand{\TrueCounterfactual}{\VarInput_{\UVWorld}^{\ConceptEdit}}
\newcommand{\ApprCounterfactual}{\VarInputAppr_{\UVWorld}^{\ConceptEdit}}
\newcommand{\TrueCounterfactualPairs}{\Pairs{\Factual}{\TrueCounterfactual}}
\newcommand{\ApprCounterfactualPairs}{\Pairs{\Factual}{\ApprCounterfactual}}
\newcommand{\SourceInput}{\VarSourceInputAbstract_{\UVEditWorld}^{\ConceptAssign}}

\newcommand{\IntDescription}{\ConceptEdit}
\newcommand{\IntDescriptionToken}{\VarToken_{\IntDescription}}

\newcommand{\SourceHidden}{\VarBigH_{\VarSourceInput}}

\newcommand{\InputHidden}{\VarBigH_{\VarInput}}
\newcommand{\HiddenForConcept}{\VarBigH^{\VarConceptAbstractSingle}}

\newcommand{\SourceHiddenForConcept}{\SourceHidden^{\VarConceptAbstractSingle}}
\newcommand{\InputHiddenForConcept}{\InputHidden^{\VarConceptAbstractSingle}}
\newcommand{\IntInvAny}[1]{\VarBigH^{#1} \IntInvSign \SourceHidden^{#1}}
\newcommand{\IntInvForConcept}{\IntInvAny{\VarConceptAbstractSingle}}
\newcommand{\ExplainerInput}{\Factual; \ConceptEdit}
\newcommand{\ExplainerTokenAppendInput}{\Factual; \IntDescriptionToken}

\newcommand{\GPTApprCounterfactualPairs}{\Pairs{\VarSourceInputAbstract_{\UVEditWorld}}{\ac{\VarSourceInputAbstract}_{\UVEditWorld}^{\ConceptEdit}}}

\usepackage{times}
\usepackage{latexsym}

\usepackage[T1]{fontenc}

\usepackage[utf8]{inputenc}

\usepackage{microtype}

\definecolor{ourRed}{HTML}{E24A33}
\definecolor{ourBlue}{HTML}{348ABD}
\definecolor{ourPurple}{HTML}{988ED5}
\definecolor{ourGray}{HTML}{777777}
\definecolor{ourLightGray}{HTML}{B8B8B8}
\definecolor{ourYellow}{HTML}{FBC15E}
\definecolor{ourGreen}{HTML}{4D8951}
\definecolor{ourPink}{HTML}{FFB5B8}
\definecolor{oursteelblue}{HTML}{9BB8D7}
\definecolor{ourOrange}{HTML}{FDBA58}
\definecolor{ourWhite}{HTML}{FAFAFA}

\newcommand{\Tabref}[1]{Table~\ref{#1}}
\newcommand{\tabref}[1]{Table~\ref{#1}}

\newcommand{\Appref}[1]{Appendix~\ref{#1}}
\newcommand{\appref}[1]{Appendix~\ref{#1}}
\newcommand{\Eqnref}[1]{Eqn.~\ref{#1}}
\newcommand{\CE}[1]{\textsc{CE}}

\newcommand{\OurMethod}{Causal Proxy Model}

\newcommand{\OurMethodAbbr}{CPM}
\newcommand{\NonInputLevelAbbr}{HI}
\newcommand{\InputLevelAbbr}{IN}
\newcommand{\nn}{\mathcal{N}}
\newcommand{\ICACEerror}{\text{ICaCE-Error}}

\newcommand{\bbm}{\mathcal{N}}

\title{Causal Proxy Models for Concept-based Model Explanations}

\author{%
Zhengxuan Wu${}^{1 \ast}$ 
,
Karel D'Oosterlink$^{1, 2 \ast}$
,
Atticus Geiger$^{1 \ast}$ 
,
Amir Zur$^{1}$ 
,
Christopher Potts$^{1}$ 
\\
\texttt{wuzhengx@cs.stanford.edu},
\texttt{karel.doosterlinck@ugent.be},
\texttt{atticusg@stanford.edu},
\\
\texttt{amirzur@stanford.edu},
\texttt{cgpotts@stanford.edu}
\\
${}^{\ast}$ Equal contribution
${}^{1}$Stanford University
${}^{2}$Ghent University
}

\iclrfinalcopy %

\begin{document}

\maketitle

\vspace{-\baselineskip}

\begin{abstract}
Explainability methods for NLP systems encounter a version of the fundamental problem of causal inference: for a given ground-truth input text, we never truly observe the counterfactual texts necessary for isolating the causal effects of model representations on outputs. In response, many explainability methods make no use of counterfactual texts, assuming they will be unavailable. In this paper, we show that robust causal explainability methods can be created using approximate counterfactuals, which can be written by humans to approximate a specific counterfactual or simply sampled using metadata-guided heuristics. The core of our proposal is the \OurMethod{} (\OurMethodAbbr{}). A \OurMethodAbbr{} explains a black-box model $\bbm{}$ because it is trained to have the same \emph{actual} input/output behavior as $\bbm{}$ while creating neural representations that can be intervened upon to simulate the \emph{counterfactual} input/output behavior of $\bbm{}$.  Furthermore, we show that the best \OurMethodAbbr{} for $\bbm{}$ performs comparably to $\bbm{}$ in making factual predictions, which means that the \OurMethodAbbr{} can simply replace $\bbm{}$, leading to more explainable deployed models. Our code is available at \url{https://github.com/frankaging/Causal-Proxy-Model}.

\end{abstract}

\section{Introduction}

The gold standard for explanation methods in AI should be to elucidate the \emph{causal role} that a model's representations play in its overall behavior -- to truly explain \emph{why} the model makes the predictions it does. Causal explanation methods seek to do this by resolving the counterfactual question of what the model would do if input $X$ were changed to a relevant counterfactual version $X'$. Unfortunately, even though neural networks are fully observed, deterministic systems, we still encounter the fundamental problem of causal inference \citep{holland1986statistics}:  for a given ground-truth input $X$, we never observe the counterfactual inputs $X'$ necessary for isolating the causal effects of model representations on outputs. The issue is especially pressing in domains where it is hard to synthesize approximate counterfactuals. In response to this, explanation methods typically do not explicitly train on counterfactuals at all.

In this paper, we show that robust explanation methods for NLP models can be obtained using texts approximating true counterfactuals. The heart of our proposal is the \OurMethod{} (\OurMethodAbbr{}). \OurMethodAbbr s are trained to mimic both the factual and counterfactual behavior of a black-box model $\bbm{}$. We explore two different methods for training such explainers. These methods share a distillation-style objective that pushes them to mimic the factual behavior of $\bbm{}$, but they differ in their counterfactual objectives. The input-based method \CPMIN{} appends to the factual input a new token associated with the counterfactual concept value. The hidden-state method $\CPMHI$ employs the Interchange Intervention Training (IIT) method of \citet{geiger-etal-2021-iit} to localize information about the target concept in specific hidden states. \Figref{fig:core-model} provides a high-level overview.

We evaluate these methods on the 
CEBaB benchmark for causal explanation methods \citep{abraham2022cebab}, which provides large numbers of original examples (restaurant reviews) with human-created counterfactuals for specific concepts (e.g., service quality), with all the texts labeled for their concept-level and text-level sentiment. We consider two types of approximate counterfactuals derived from CEBaB: texts written by humans to approximate a specific counterfactual, and texts sampled using metadata-guided heuristics. Both approximate counterfactual strategies lead to state-of-the-art performance on CEBaB for both \CPMIN{} and \CPMHI. 

We additionally identify two other benefits of using \OurMethodAbbr s to explain models. First, both \CPMIN{} and \CPMHI{} have factual performance comparable to that of the original black-box model $\bbm{}$ and can explain their own behavior extremely well. Thus, the \OurMethodAbbr{} for $\bbm{}$ can actually replace $\bbm{}$, leading to more explainable deployed models. Second, \CPMHI{} models localize concept-level information in their hidden representations, which makes their behavior on specific inputs very easy to explain. 
We illustrate this using Path Integrated Gradients \citep{sundararajan17a}, which we adapt to allow input-level attributions to be mediated by the intermediate states that were targeted for localization.
Thus, while both \CPMIN{} and \CPMHI{} are comparable as explanation methods according to CEBaB, the qualitative insights afforded by \CPMHI{} models may given them the edge when it comes to explanations.

\section{Related Work}

Understanding model behavior serves many goals for large-scale AI systems, including transparency~\citep{kim2015,lipton_mythos_2018,pearl_limitations_2019,ehsan2021expanding}, trustworthiness~\citep{ribeiro_why_2016,guidotti2018survey,jacovi2020towards,jakesch2019ai}, safety~\citep{amodei_concrete_2016,otte_safe_2013}, and fairness~\citep{hardt_equality_2016, kleinberg2017,goodman_european_2017,mehrabi2021survey}. With \OurMethodAbbr s, our goal is to achieve explanations that are causally motivated and concept-based, and so we concentrate here on relating existing methods to these two goals.

Feature attribution methods estimate the importance of features, generally by inspecting learned weights directly or by perturbing features and studying the effects this has on model behavior~\citep{molnar2020interpretable,ribeiro_why_2016}. Gradient-based feature attribution methods extend this general mode of explanation to the hidden representations in deep networks \citep{Zeiler2014,springerberg2014,Binder16,Shrikumar16,sundararajan17a}. Concept Activation Vectors (CAVs; \citealt{kim_interpretability_2018, yeh2020completeness}) can also be considered feature attribution methods, as they probe for semantically meaningful directions in the model's internal representations and use these to estimate the importance of concepts on the model predictions. While some methods in this space do have causal interpretations (e.g., \citealt{sundararajan17a,yeh2020completeness}), most do not. In addition, most of these methods offer explanations in terms of specific (sets of) features/neurons. (Methods based on CAVs operate directly in terms of more abstract concepts.)

Intervention-based methods study model representations by modifying them in systematic ways and observing the resulting model behavior. These methods are generally causally motivated and allow for concept-based explanations. Examples of methods in this space include causal mediation analysis \citep{vig2020causal, de2021sparse, ban2022testing}, causal effect estimation \citep{feder_causalm_2020,elazar2021amnesic,abraham2022cebab,lovering2022unit}, tensor product decomposition \citep{soulos-etal-2020-discovering}, and causal abstraction analysis \citep{geiger-etal-2020-neural, geiger2021causal}. \OurMethodAbbr s are most closely related to the method of IIT \citep{geiger2021causal}, which extends causal abstraction analysis to optimization.

Probing is another important class of explanation method. Traditional probes do not intervene on the target model, but rather only seek to find information in it via supervised models \citep{conneau-etal-2018-cram,tenney-etal-2019-bert} or unsupervised models \citep{clark-etal-2019-bert,Manning-etal:2020,saphra-lopez-2019-understanding}. Probes can identify concept-based information, but they cannot offer guarantees that probed information is relevant for model behavior \citep{geiger2021causal}. For causal guarantees, it is likely that some kind of intervention is required. For example, \citet{elazar2021amnesic} and \citet{feder_causalm_2020} remove information from model representations to estimate the causal role of that information. Our \OurMethodAbbr s employ a similar set of guiding ideas but are not limited to removing information.

Counterfactual explanation methods aim to explain model behavior by providing a counterfactual example that changes the model behavior \citep{goyal2019counterfactual, verma2020counterfactual, wu2021polyjuice}. Counterfactual explanation methods are inherently causal. If they can provide counterfactual examples with regard to specific concepts, they are also concept-based.

Some explanation methods train a model making explicit use of intermediate variables representing concepts. Manipulating these intermediate variables at inference time yields causal concept-based model explanations \citep{koh2020concept,kunzel2019metalearners}. 

Evaluating methods in this space has been a persistent challenge. In prior literature, explanation methods have often been evaluated against synthetic datasets~\citep{feder_causalm_2020, yeh2020completeness}. In response, \citet{abraham2022cebab} introduced the CEBaB dataset, which provides a human-validated concept-based dataset to truthfully evaluate different causal concept-based model explanation methods. Our primary evaluations are conducted on CEBaB.

\section{\OurMethod{} (\OurMethodAbbr{})}\label{sec:CPM}

\input{figures/MainMethod-v2}

\OurMethod s (\OurMethodAbbr s) are causal concept-based explanation methods. Given a factual input $\Factual$ and a description of a concept intervention $\ConceptEdit$, they estimate the effect of the intervention on model output. The present section introduces our two core \OurMethodAbbr{} variants in detail. We concentrate here on introducing the structure of these models and their objectives, and we save discussion of associated metrics for explanation methods for \Secref{sec:cpm-eval}.

\paragraph{A Structural Causal Model}

Our discussion is grounded in the causal model depicted in \figref{fig:datagen}, which aligns well with the CEBaB benchmark. Two exogenous variables $\VarBigU$ and $\VarBigV$ together represent the complete state of the world and generate some textual data $\VarBigX$. The effect of exogenous variable $\VarBigU$ on the data $\VarBigX$ is completely mediated by a set of intermediate variables $\VarConceptAbstract_1, \VarConceptAbstract_2 \dots, \VarConceptAbstract_k$, which we refer to as \textit{concepts}. Therefore, we can think of $\VarBigU$ as the part of the world that gives rise to these concepts $\{\VarConceptAbstract\}_1^k$.

Using this causal model, we can describe counterfactual data -- data that arose under a counterfactual state of the world (right diagram in \figref{fig:datagen}). Our factual text is $\Factual$, and we use $\TrueCounterfactual$ for the counterfactual text obtained by intervening on concept $\VarConceptAbstractSingle$ to set its value to $\VarConceptEdit$. The counterfactual $\TrueCounterfactual$ describes the output when the value of $\VarConceptAbstractSingle$ is set to $\VarConceptEdit$, all else being held equal.

\paragraph{Approximate Counterfactuals}\label{sec:approx-strategies}

Unfortunately, pairs like $\TrueCounterfactualPairs$ are never observed, and thus we need strategies for creating approximate counterfactuals $\ApprCounterfactual$. \Figref{fig:approx-strategies} describes the two strategies we use in this paper. In the human-created strategy, we rely on a crowdworker to edit $\Factual$ to achieve a particular counterfactual goal -- say, making the evaluation of the restaurant's food negative. CEBaB contains an abundance of such pairs $\ApprCounterfactualPairs$. However, CEBaB is unusual in having so many human-created approximate counterfactuals, so we also explore a simpler strategy in which $\ApprCounterfactual$ is sampled with the requirement that it match $\Factual$ on all concepts but sets $\VarConceptAbstractSingle$ to $\VarConceptEdit$. This strategy is supported in many real-world datasets -- for example, the OpenTable reviews underlying CEBaB all have the needed metadata~\citep{abraham2022cebab}.

\paragraph{\CPMIN: Input-based \OurMethodAbbr{}} \label{subsubsec:cpm-in}

To train $\CPMIN$, we associate a new randomly initialized token $\IntDescriptionToken$ with each unique intervention description $\IntDescription$. Given a dataset of approximate counterfactual pairs $\ApprCounterfactualPairs$ and a black-box model $\bbm{}$, we train a new $\CPMIN$ model $\mathcal{P}$ under a weighted sum of the following objectives:
\begin{align}
    \mathcal{L}_{\text{Mimic}} &= 
    \text{CE}_{\text{S}}\big(\bbm{}(\Factual), \mathcal{P}(\Factual)\big) 
    \label{eq:cpm-in:factual}
    \\
     \mathcal{L}_{\text{IN}} &= 
     \text{CE}_{\text{S}}\big(\bbm{}(\ApprCounterfactual), \mathcal{P}(\Factual ; \IntDescriptionToken)\big) 
     \label{eq:cpm-in:counterfactual}
\end{align}
where $\Factual; \IntDescriptionToken$ in \Eqnref{eq:cpm-in:counterfactual} denotes the concatenation of the factual input and the token describing the intervention. $\text{CE}_{\text{S}}$ represents the smoothed cross-entropy loss \citep{hinton2015distilling}, measuring the divergence between the output logits of both models. The objective in \Eqnref{eq:cpm-in:factual} pushes $\mathcal{P}$ to predict the same output as $\bbm{}$ under conventional circumstances (\figref{fig:l-mimic}), while \Eqnref{eq:cpm-in:counterfactual} pushes $\mathcal{P}$ to predict the counterfactual behavior of $\bbm{}$ when a descriptor of the intervention is given (\figref{fig:l-cpm-in}).\footnote{These objectives are described with regard to a single approximate counterfactual pair for the sake of clarity. At train-time, we aggregate the objective over all considered training pairs. We take $\VarConceptAbstractSingle{}$ to always represent the intervened-upon concept. The weights of $\bbm{}$ are frozen. } 

At inference time, approximate counterfactuals are inaccessible. To explain model $\bbm{}$, we append the newly learned descriptor tokens $\IntDescriptionToken$ to a factual input, upon which $\mathcal{P}$ predicts a counterfactual output for this input, used to estimate the counterfactual behavior of $\bbm{}$ under this intervention.

\paragraph{\CPMHI: Hidden-state \OurMethodAbbr{}} \label{subsubsec:cpm-hi}
 
Our $\CPMHI$ models are trained on the same data and with the same set of goals as $\CPMIN$, to mimic both the factual and counterfactual behavior of $\bbm{}$. The key difference is how the information about the intervention $\IntDescription$ is exposed to the model. Specifically, we adapt Interchange Intervention Training ~\citep{geiger-etal-2021-iit} to train our $\CPMHI$ models for concept-based model explanation.

A conventional intervention on a hidden representation $\VarBigH$ of a neural network $\bbm{}$ fixes the value of the representation $\VarBigH$ to a constant. In an interchange intervention, we instead fix $\VarBigH$ to the value it would have been when processing a separate source input $\VarSourceInput$. The result of the interchange intervention is a new model. Formally, we describe this new model as $\nn_{\IntInvAny{}}$, where $\leftarrow$ is the conventional intervention operator and $\SourceHidden$ is the value of hidden representation $H$ when processing input $\VarSourceInput$.

Given a dataset of approximate counterfactual input pairs $\ApprCounterfactualPairs$ and a black-box model $\bbm{}$, we train a new $\CPMHI$ model $\mathcal{P}$ under a weighted sum of the previous factual objective (\Eqnref{eq:cpm-in:factual}) and the following counterfactual objective:
\begin{align}
     \mathcal{L}_{\text{HI}} &= 
     \text{CE}_{\text{S}}\big(\bbm{}(\ApprCounterfactual), \mathcal{P}_{\IntInvForConcept}(\Factual)\big) 
     \label{eq:cpm-hi:counterfactual}
\end{align}
Here $\HiddenForConcept$ are hidden states designated for concept $\VarConceptAbstractSingle$. In essence, we train $\mathcal{P}$ to fully mediate the effect of intervening on $\VarConceptAbstractSingle$ in the hidden representation $\HiddenForConcept$. The source input $\VarSourceInput$ is any input $\SourceInput$  that has $\ConceptAssign$. As $\mathcal{P}$ only receives information about the concept-level intervention $\IntDescription$ via the interchange intervention $\IntInvForConcept$, the model is forced to store all causally relevant information with regard to $\VarConceptAbstractSingle$ in the corresponding hidden representation. This process is described in \figref{fig:l-cpm-hi}. 

In the ideal situation, the source input $\SourceInput$ and $\Factual$ share the same value only for $\VarConceptAbstractSingle$ and differ on all others, so that the counterfactual signal needed for localization is pure. However, we do not insist on this when we sample. In addition, we allow \emph{null effect} pairs in which $\Factual$ and $\ApprCounterfactual$ are identical. For additional details on this sampling procedure, see \Appref{app:pairs-sampling}. 

At inference time, approximate counterfactuals are inaccessible, as before. To explain model $\bbm{}$ with regard to intervention $\IntDescription$, we manipulate the internal states of model $\mathcal{P}$ by intervening on the localized representation $\HiddenForConcept$ for concept $\VarConceptAbstractSingle$. To achieve this, we sample a source input $\SourceInput$ from the train set as any input $\VarInput$ that has $\ConceptAssign$ to derive $\SourceHiddenForConcept$.

\section{Experiment Setup}\label{sec:cpm-eval}

\subsection{Causal Estimation-Based Benchmark (CEBaB)}\label{subsec:cebab}

CEBaB \citep{abraham2022cebab} is a large benchmark of high-quality, labeled approximate counterfactuals for the task of sentiment analysis on restaurant reviews. The benchmark was created starting from a set of 2,299 original restaurant reviews from OpenTable. For each of these original reviews, approximate counterfactual examples were written by human annotators; the annotators were tasked to edit the original text to reflect a specific intervention, like `change the food evaluation from negative to positive' or `change the service evaluation from positive to unknown'. In this way, the original reviews were expanded with approximate counterfactuals to a total of 15,089 texts. The groups of originals and corresponding approximate counterfactuals are partitioned over train, dev, and test sets. The pairs in the development and test set are used to benchmark explanation methods. 

Each text in CEBaB was labeled by five crowdworkers with a 5-star sentiment score. In addition, each text was annotated at the concept level for four mediating concepts $\{\VarConceptAbstract_{\text{ambiance}}$, $\VarConceptAbstract_{\text{food}}$, $\VarConceptAbstract_{\text{noise}}$, and $\VarConceptAbstract_{\text{service}}\}$, using the labels $\{\text{negative}, \text{unknown}, \text{positive}\}$, again with five crowdworkers annotating each concept-level label. We refer to \appref{app:dataset-stats} and \citealt{abraham2022cebab} for additional details.

As discussed above (\secref{sec:approx-strategies} and \figref{fig:approx-strategies}), we consider two sources of approximate counterfactuals using CEBaB. For human-created counterfactuals, we use the edited restaurant reviews of the train set. For metadata-sampled counterfactuals, we  sample factual inputs from the train set that have the desired combination of mediating concepts. Using all the human-created edits leads to 19,684 training pairs of factuals and corresponding approximate counterfactuals. Sampling counterfactuals leads to 74,574 pairs. We use these approximate counterfactuals to train explanation methods. \Appref{app:pairs-sampling} provides more information about our pairing process.

\subsection{Evaluation Metrics}

Much of the value of a benchmark like CEBaB derives from its support for directly calculating the Estimated Individual Causal Concept Effect ($\widehat{\text{ICaCE}}_{\bbm{}}$) for a model $\bbm{}$ given a human-generated approximate counterfactual pair $\ApprCounterfactualPairs$:
\begin{equation}\label{eq:icace}
\widehat{\text{ICaCE}}_{\nn}\ApprCounterfactualPairs = 
\bbm{}(\ApprCounterfactual) - \bbm{}(\Factual)
\end{equation}
This is simply the difference between the vectors of output scores for the two examples.

We do not expect to have pairs $\ApprCounterfactualPairs$ at inference time, and this is what drives the development of explanation methods $\mathcal{E}_{\bbm{}}$ that \textit{estimate} this quantity using only a factual input $\Factual$ and a description of the intervention $\IntDescription$. To benchmark such methods, we follow \citet{abraham2022cebab} in using the $\ICACEerror$:
\begin{equation}\label{eq:ICaCE-Error}
    \ICACEerror_{\nn}^{\mathcal{D}}(\mathcal{E}) = 
    \frac{1}{\left|\mathcal{D}\right|} 
    \negthickspace
    \sum_{\ApprCounterfactualPairs \in \mathcal{D}}
    \negthickspace 
    \! \! \! \!\mathsf{Dist}\big(
    \widehat{\text{ICaCE}}_{\nn}(\ApprCounterfactualPairs), 
    \mathcal{E}_{\nn}(\ExplainerInput)
    \big)
\end{equation}
Here, we assume that $\mathcal{D}$ is a dataset consisting entirely of approximate counterfactual pairs $\ApprCounterfactualPairs$. $\mathsf{Dist}$ measures the distance between the $\widehat{\text{ICaCE}}_{\bbm{}}$ for the model $\bbm{}$ and the effect predicted by the explanation method. \citet{abraham2022cebab} consider three values for $\mathsf{Dist}$: L2, which captures both direction and magnitude; Cosine distance, which captures the direction of effects but not their magnitude; and NormDiff (absolute difference of L2 norms), which captures magnitude but not direction. We report all three metrics.

\subsection{Baseline Methods} \label{subsec:baselines}

\paragraph{$\text{BEST}_\text{CEBaB}$}
We compare our results with the best results obtained on the CEBaB benchmark. Crucially, $\text{BEST}_\text{CEBaB}$ is not a single method, but rather pools together the best result obtained by \textit{any} explanation method previously benchmarked on the CEBaB dataset, for every  combination of model and metric.

\paragraph{S-Learner}
Our version of S-Learner \citep{kunzel2019metalearners} learns to mimic the factual behavior of black-box model $\bbm{}$ while making the intermediate concepts explicit.\footnote{We use the finetuned concept-level sentiment analysis models released by  \citet{abraham2022cebab}.} Given a factual input, a finetuned BERT model $\mathcal{B}$ first predicts values for the intermediate concepts. Then, a logistic regression model $\mathsf{LR}_{\bbm{}}$ is trained to map these intermediate concept values to the factual output of black-box model $\bbm{}$, under the following objective.\footnote{We use the default implementation \texttt{LogisticRegression} of \texttt{scikit-learn} \citep{sklearn_api}.}
\begin{equation}
\mathcal{L}_{\text{Mimic}}^{\text{S}, \mathcal{B}} =     \text{CE}_{\text{S}}\big(\bbm{}(\Factual), \mathsf{LR}_{\bbm{}}(\mathcal{B}(\Factual))\big) 
\end{equation}
By intervening on the intermediate predicted concept values at inference-time, we can hope to simulate the counterfactual behavior of $\bbm{}$:
\begin{equation}
\mathcal{E}_{\nn}^{\text{S}, \mathcal{B}}(\ExplainerInput) 
= 
\mathsf{LR}_{\bbm{}}((\mathcal{B}(\Factual))_{\IntDescription}) 
- 
\mathsf{LR}_{\bbm{}}(\mathcal{B}(\Factual))
\end{equation}
When using S-Learner in conjunction with approximate counterfactual inputs at train-time, we simply add this counterfactual data on top of the observational data that is typically used to train S-Learner.

\paragraph{GPT-3}
Large language models such as \texttt{GPT-3} (175B) have shown extraordinary power in terms of in-context learning~\citep{brown_language_2020}.\footnote{We use the largest \texttt{davinci} model publicly available at \url{https://beta.openai.com/playground}.} We use \texttt{GPT-3} to generate a new approximate counterfactual at inference time given a factual input and a descriptor of the intervention. This generated counterfactual is directly used to estimate the change in model behavior:
\begin{equation}
\mathcal{E}_{\nn}^{\text{GPT-3}}(\ExplainerInput) 
= 
\bbm{}(\texttt{GPT-3}(\Factual ; \IntDescription)) 
- 
\bbm{}(\Factual)
\end{equation}

We use our train-time approximate counterfactual inputs to construct a prompt for \texttt{GPT-3}. Given this prompt, \texttt{GPT-3} outputs new approximate counterfactuals given a factual and intervention descriptor. Full details on how these prompts are constructed can be found in \appref{app:gpt-3}.

\subsection{Causal Proxy Models} \label{subsec:cpm-details}

We train \OurMethodAbbr s for the publicly available models released for CEBaB, fine-tuned as five-way sentiment classifiers on the factual data. This includes four model architectures: \texttt{bert-base-uncased} (\texttt{BERT}; \citealt{devlin_bert_2019}), \texttt{RoBERTa-base} (\texttt{RoBERTa}; \citealt{liu_roberta_2019}), \texttt{GPT-2} (\texttt{GPT-2}; \citealt{radford2019language}), and \texttt{LSTM+GloVe} (\texttt{LSTM}; \citealt{Hochreiter:Schmidhuber:1997,pennington-etal-2014-glove}). All Transformer-based models \citep{Vaswani-etal:2017} have 12 Transformer layers. Before training, each \OurMethodAbbr{} model is initialized with the architecture and weights of the black-box model we aim to explain. Thus, the \OurMethodAbbr s are rooted in the factual behavior of $\bbm{}$ from the start. We include details about our setup in \appref{app:training-setups}.

The inference time comparisons for these models are as follows, where $\mathcal{P}$ in \eqref{eq:cpm_in_estimate} and \eqref{eq:cpm_hi_estimate} refers to the \OurMethodAbbr{} model trained under \CPMIN{} and \CPMHI{} objectives, respectively:
\begin{align}
\mathcal{E}_{\nn}^{\CPMIN}(\ExplainerInput) 
&= 
\mathcal{P}(\ExplainerTokenAppendInput) 
- 
\mathcal{N}(\Factual)
\label{eq:cpm_in_estimate}
\\
\mathcal{E}_{\nn}^{\CPMHI}(\ExplainerInput) 
&= 
\mathcal{P}_{\IntInvForConcept}(\Factual) 
- 
\mathcal{N}(\Factual) 
\label{eq:cpm_hi_estimate}
\end{align}
Here, $\VarSourceInput$ is a source input with $\ConceptAssign$, and $\HiddenForConcept$ is the neural representation associated with $\VarConceptAbstractSingle$ which takes value $\SourceHiddenForConcept$ on the source input $\VarSourceInput$. As $\HiddenForConcept$, for \texttt{BERT} we use slices of width 192 taken from the 1st intermediate token of the 10th layer. For \texttt{RoBERTa}, we use the 8th layer instead. For \texttt{GPT-2}, we pick the final token of the 12th layer, again with slice width of 192. For \texttt{LSTM}, we consider slices of the attention-gated sentence embedding with width 64. \Appref{sec:iit-location} studies the impact of intervention location and size.

Following the guidance on IIT given by \citet{geiger-etal-2021-iit}, we train \CPMHI{} with an additional multi-task objective as,
\begin{equation}\label{eq:loss-multi}
        \mathcal{L}_{\text{Multi}} = \sum_{\VarConceptAbstractSingle \in \VarConceptAbstract}\text{CE}(\mathsf{MLP}(\InputHiddenForConcept), \VarConcept)
\end{equation}
where our probe is parameterized by a multilayer perceptron $\mathsf{MLP}$, and $\InputHiddenForConcept$ is the value of hidden representation for the concept $\VarConceptAbstractSingle$ when processing input $\VarInput$ with a concept label of $\VarConcept$ for $\VarConceptAbstractSingle$.

\section{Results}

\newcommand{\MetricColName}{Metric}
\newcommand{\MetricL}{L2} %
\newcommand{\MetricCos}{Cosine} %
\newcommand{\MetricNormDiff}{NormDiff} %
\newcommand{\MetricF}{$\texttt{Macro-F1}$}
\newcommand{\ci}[1]{{\scriptsize(\gobblechar[v]{#1})}}
\newcommand{\BoldHuman}[1]{\underline{\textbf{#1}}}
\newcommand{\BoldSampling}[1]{\textbf{#1}}
\newcommand{\BoldGlobal}[1]{\textbf{#1}}

\begin{table*}[tp]
\centering
\resizebox{1.00\linewidth}{!}{%
  \centering
  \setlength{\tabcolsep}{4pt}
  \begin{tabular}{@{} l@{ } r ccccccccccccc @{}}
\toprule
\multirow{3}{*}{} 
&
&
& \multicolumn{2}{c}{\emph{no counterfactuals}} 
& & \multicolumn{4}{c}{\emph{sampled counterfactuals}} 
& & \multicolumn{4}{c}{\emph{human-created counterfactuals}} 
\\
&  
& & & & & & &
\multicolumn{1}{l}{(\textbf{ours})} &
\multicolumn{1}{l}{(\textbf{ours})}
& & & &
\multicolumn{1}{l}{(\textbf{ours})} &
\multicolumn{1}{l}{(\textbf{ours})}
\\
Model & \MetricColName 
&
& \multirow{1}{*}{BEST$_{\text{CEBaB}}$} 
& \multicolumn{1}{c}{\texttt{S-Learner}} 
& 
& \multicolumn{1}{c}{\texttt{S-Learner}} 
& \multicolumn{1}{c}{\texttt{GPT-3}} %
& \multicolumn{1}{c}{\textbf{\CPMIN}} %
& \multicolumn{1}{c}{\textbf{\CPMHI}} %
&
& \multicolumn{1}{c}{\texttt{S-Learner}} 
& \multicolumn{1}{c}{\texttt{GPT-3}} %
& \multicolumn{1}{c}{\textbf{\CPMIN}} %
& \multicolumn{1}{c}{\textbf{\CPMHI}} %
\\
\cmidrule{1-2} \cmidrule{4-5} \cmidrule{7-10} \cmidrule{12-15}
\multirow{3}{*}{\texttt{BERT}} 
& \MetricL 
&
& 0.74 \ci{0.02} & 0.74 \ci{0.02} 
&
& 0.74 \ci{0.02} & 0.71 \ci{0.01} & 0.63 \ci{0.01} & \BoldGlobal{0.60} \ci{0.01} 
&  
& 0.73 \ci{0.02} & \BoldGlobal{0.45} \ci{0.01} & \BoldGlobal{0.45} \ci{0.02} & \BoldGlobal{0.45} \ci{0.03} \\
& \MetricCos 
& 
& 0.59 \ci{0.03} & 0.63 \ci{0.01} 
&  
& 0.63 \ci{0.01} & 0.51 \ci{0.00} & 0.46 \ci{0.00} & \BoldGlobal{0.45} \ci{0.00} 
&  
& 0.60 \ci{0.01} & 0.36 \ci{0.00} & \BoldGlobal{0.35} \ci{0.00} & 0.36 \ci{0.04} \\
& \MetricNormDiff 
&
& 0.44 \ci{0.01} & 0.54 \ci{0.02} 
&  
& 0.53 \ci{0.02} & \BoldGlobal{0.35} \ci{0.01} & 0.39 \ci{0.01} & {0.38}  \ci{0.00} 
&  
& 0.52 \ci{0.02} & 0.25 \ci{0.00} & \BoldGlobal{0.24} \ci{0.01} & 0.27 \ci{0.01} \\
\cmidrule{1-2} \cmidrule{4-5} \cmidrule{7-10} \cmidrule{12-15}
\multirow{3}{*}{\texttt{RoBERTa}} 
& \MetricL 
&
& 0.78 \ci{0.01} & 0.78 \ci{0.01} 
&  
& 0.78 \ci{0.00} & 0.74 \ci{0.01} & \BoldGlobal{0.66} \ci{0.01} & {0.67} \ci{0.02} 
&  
& 0.77 \ci{0.00} & 0.48 \ci{0.01} & \BoldGlobal{0.46} \ci{0.01} & 0.47 \ci{0.03} \\
& \MetricCos
&
& 0.58 \ci{0.01} & 0.64 \ci{0.01} 
&  
& 0.65 \ci{0.01} & 0.53 \ci{0.01} & \BoldGlobal{0.46} \ci{0.00} & {0.47} \ci{0.00} 
&  
& 0.63 \ci{0.01} & 0.39 \ci{0.00} & \BoldGlobal{0.38} \ci{0.01} & 0.39 \ci{0.03} \\
& \MetricNormDiff 
&
& {0.45} \ci{0.00} & 0.59 \ci{0.01} 
&  
& 0.58 \ci{0.00} & \BoldGlobal{0.36} \ci{0.00} & 0.42 \ci{0.01} & {0.45} \ci{0.03} 
&  
& 0.56 \ci{0.00} & 0.28 \ci{0.01} & \BoldGlobal{0.26} \ci{0.01} & 0.29 \ci{0.05} \\
\cmidrule{1-2} \cmidrule{4-5} \cmidrule{7-10} \cmidrule{12-15}
\multirow{3}{*}{\texttt{GPT-2}} 
& \MetricL 
&
& 0.60 \ci{0.02} & 0.60 \ci{0.02} 
&  
& 0.61 \ci{0.01} & 0.65 \ci{0.01} & 0.55 \ci{0.01} & \BoldGlobal{0.51} \ci{0.01} 
&  
& 0.61 \ci{0.01} & 0.43 \ci{0.01} & \BoldGlobal{0.41} \ci{0.01} & \BoldGlobal{0.41} \ci{0.04} \\
& \MetricCos 
&
& 0.59 \ci{0.01} & 0.59 \ci{0.01} 
&  
& 0.59 \ci{0.01} & 0.52 \ci{0.00} & 0.47 \ci{0.01} & \BoldGlobal{0.46} \ci{0.00} 
&  
& 0.59 \ci{0.01} & 0.40 \ci{0.00} & \BoldGlobal{0.37} \ci{0.01} & 0.39 \ci{0.05} \\
& \MetricNormDiff 
&
& 0.40 \ci{0.01} & 0.40 \ci{0.01} 
&  
& 0.41 \ci{0.01} & 0.34 \ci{0.00} & 0.32 \ci{0.01} & \BoldGlobal{0.30} \ci{0.00} 
&  
& 0.40 \ci{0.01} & 0.24 \ci{0.01} & \BoldGlobal{0.23} \ci{0.01} & 0.27 \ci{0.05} \\
\cmidrule{1-2} \cmidrule{4-5} \cmidrule{7-10} \cmidrule{12-15}
\multirow{3}{*}{\texttt{LSTM}} 
& \MetricL 
&
& 0.73 \ci{0.01} & 0.73 \ci{0.01} 
&  
& 0.73 \ci{0.01} & 0.76 \ci{0.00} & 0.66 \ci{0.01} & \BoldGlobal{0.64} \ci{0.02} 
&  
& 0.72 \ci{0.00} & \BoldGlobal{0.49} \ci{0.00} & 0.52 \ci{0.00} & 0.54 \ci{0.01} \\
& \MetricCos 
&
& 0.64 \ci{0.01} & 0.64 \ci{0.01} 
& 
& 0.64 \ci{0.01} & 0.57 \ci{0.01} & \BoldGlobal{0.50} \ci{0.00} & \BoldGlobal{0.50} \ci{0.01} 
&  
& 0.63 \ci{0.01} & \BoldGlobal{0.44} \ci{0.00} & 0.45 \ci{0.01} & 0.46 \ci{0.00} \\
& \MetricNormDiff 
&
& 0.50 \ci{0.01} & 0.53 \ci{0.01} 
&  
& 0.53 \ci{0.00} & \BoldGlobal{0.41} \ci{0.00} & 0.42 \ci{0.00} & \BoldGlobal{0.41} \ci{0.01} 
&  
& 0.54 \ci{0.00} & \BoldGlobal{0.30} \ci{0.00} & 0.34 \ci{0.01} & 0.36 \ci{0.00} \\
\bottomrule
\end{tabular}}
  \caption{CEBaB scores measured in three different metrics on the test set for four different model architectures as a five-class sentiment classification task. \textbf{Lower is better}. Results averaged over three distinct seeds, standard deviations in parentheses. The metrics are described in~\secref{sec:cpm-eval}. Best averaged result is bolded (including ties) per approximate counterfactual creation strategy.}
  \label{tab:cebab-main}
\end{table*}

We first benchmark both  \OurMethodAbbr{} variants and our baseline methods on CEBaB. We show that the \OurMethodAbbr s achieve state-of-the-art performance, for both types of approximate counterfactuals used during training (\secref{sec:result-cebab}). Given the good factual performance achieved by \OurMethodAbbr s, we subsequently investigate whether \OurMethodAbbr s can be deployed both as predictor and explanation method at the same time (\secref{sec:self-explain}) and find that they can. Finally, we show that the localized representations of \CPMHI{} give rise to concept-aware feature attributions (\secref{sec:ig}). Our supplementary materials report on detailed ablation studies and explore the potential of our methods for model debiasing.

\subsection{CEBaB Performance} \label{sec:result-cebab}

\Tabref{tab:cebab-main} presents our main results. The results are grouped per approximate counterfactual type used during training. Both \CPMIN{} and \CPMHI{} beat BEST$_{\text{CEBaB}}$ in every evaluation setting by a large margin, establishing state-of-the-art explanation performance. Interestingly, \CPMHI{} seems to slightly outperform \CPMIN{} using \emph{sampled} approximate counterfactuals, while slightly underperforming \CPMIN{} on \emph{human-created} approximate counterfactuals.
\Appref{sec:iit-loss-terms} reports on ablation studies that indicate that, for \CPMHI, this state-of-the-art performance is primarily driven by the role of IIT in localizing concepts.

\texttt{S-Learner}, one of the best individual explainers from the original CEBaB paper \citep{abraham2022cebab}, shows only a marginal improvement when naively incorporating \emph{sampled} and \emph{human-created} counterfactuals during training over using \emph{no counterfactuals}. This indicates that the large performance gains achieved by our \OurMethodAbbr s over previous explainers are most likely due to the explicit use of a counterfactual training signal, and not primarily due to the addition of extra (counterfactual) data.

\texttt{GPT-3} occasionally performs on-par with our \OurMethodAbbr s, generally only slightly underperforming our best explainer on \emph{human-created counterfactuals}, while being significantly worse on \emph{sampled counterfactuals}. While the \texttt{GPT-3} explainer also explicitly uses approximate counterfactual data, the results indicate that our proposed counterfactual mimic objectives give better results. The better performance of \OurMethodAbbr s when considering \textit{sampled counterfactuals} over \texttt{GPT-3} shows that our approach is more robust to the quality of the approximate counterfactuals used. While the \texttt{GPT-3} explainer is easy to set up (no training required), it might not be suitable for some explanation applications regardless of performance, due to the latency and cost involved in querying the \texttt{GPT-3} API.

Across the board, explainers trained with \emph{human-created} counterfactuals are better than those trained with \emph{sampled} counterfactuals. This shows that the performance of explanation methods depends on the quality of the approximate counterfactual training data. While human counterfactuals give excellent performance, they may be expensive to create. Sampled counterfactuals are cheaper if the relevant metadata is available. Thus, under budgetary constraints, sampled counterfactuals may be more efficient.

Finally, \CPMIN{} is conceptually the simpler of the two \OurMethodAbbr{} variants. However, we discuss in \secref{sec:ig} how the localized representations of \CPMHI{} lead to additional explainability benefits.

\subsection{Self-Explanation with \OurMethodAbbr{}}\label{sec:self-explain}

As outlined in \secref{sec:CPM}, \OurMethodAbbr s learn to mimic both the factual and counterfactual behavior of the black-box models they are explaining. We show in \tabref{tab:self-explain-baselines-mf1} that our \OurMethodAbbr s achieve a factual \MetricF{} score comparable to the black-box finetuned models.

We investigate if we can simply replace the black-box model with our \OurMethodAbbr{} and use the \OurMethodAbbr{} both as factual predictor and counterfactual explainer. To answer this questions, we measure the self-explanation performance of \OurMethodAbbr s by simply replacing the black-box model $\bbm{}$ in \Eqnref{eq:ICaCE-Error} with our factual \OurMethodAbbr{} predictions at inference time.  

\Tabref{tab:self-explain-baselines} reports these results. We find that both \CPMIN{} and \CPMHI{} achieve better self-explanation performance compared to providing explanations for another black-box model. Furthermore, \CPMHI{} provides better self-explanation than \CPMIN{}, suggesting our interchange intervention procedure leads the model to localize concept-based information in hidden representations. This shows that \OurMethodAbbr s may be viable as replacements for their black-box counterpart, since they provide similar task performance while providing faithful counterfactual explanations of both the black-box model and themselves.

\begin{figure}[!t]
\begin{minipage}{\textwidth}

\begin{minipage}[b]{0.48\textwidth}
\centering
\resizebox{\linewidth}{!}{%
      \centering
      \setlength{\tabcolsep}{3pt}
      \begin{tabular}[b]{@{} l@{ \ } cccccccc @{}}
        \toprule
        & & & & \multicolumn{2}{c}{\emph{sampled}} & & \multicolumn{2}{c}{\emph{human-created}} \\
        & & Fine- & & \multicolumn{2}{c}{\emph{counterfactuals}} & & \multicolumn{2}{c}{\emph{counterfactuals}} \\
        \multirow{1}{*}{Model} & & \multirow{1}{*}{tuned} & & \multicolumn{1}{c}{\textbf{\CPMIN}} & \multicolumn{1}{c}{\textbf{\CPMHI}} & & \multicolumn{1}{c}{\textbf{\CPMIN}} & \multicolumn{1}{c}{\textbf{\CPMHI}} \\
        \cmidrule{1-1} \cmidrule{3-3} \cmidrule{5-6} \cmidrule{8-9}
        \texttt{BERT}
        & & 0.70 \ci{0.01} &  & 0.70 \ci{0.00} & 0.67 \ci{0.02} &  & 0.70 \ci{0.01} & 0.69 \ci{0.01} \\
        \texttt{RoBERTa}
        & & 0.70 \ci{0.00} &  & 0.70 \ci{0.00} & 0.69 \ci{0.01} &  & 0.71 \ci{0.01} & 0.71 \ci{0.00} \\
        \texttt{GPT-2}
        & & 0.65 \ci{0.00} &  & 0.65 \ci{0.00} & 0.67 \ci{0.01} &  & 0.66 \ci{0.01} & 0.68 \ci{0.00} \\
        \texttt{LSTM}
        & & 0.60 \ci{0.01} &  & 0.60 \ci{0.01} & 0.56 \ci{0.00} &  & 0.54 \ci{0.00} & 0.59 \ci{0.01} \\
        \bottomrule
        \end{tabular}
    }
  \captionof{table}{Task performance measured as \texttt{Macro-F1} score on the test set. Results averaged over three distinct seeds; standard deviations in parentheses.}
  \label{tab:self-explain-baselines-mf1}
\end{minipage}
\hfill
\begin{minipage}[b]{0.48\textwidth}
\centering
\resizebox{\linewidth}{!}{%
      \centering
      \setlength{\tabcolsep}{4pt}
      \begin{tabular}[b]{@{} l@{}r cccccc @{}}
        \toprule
        & & & \multicolumn{2}{c}{\emph{sampled}} & & \multicolumn{2}{c}{\emph{human-created}} \\
        & & & \multicolumn{2}{c}{\emph{counterfactuals}} & & \multicolumn{2}{c}{\emph{counterfactuals}} \\
        \multirow{1}{*}{Model} & \multirow{1}{*}{\MetricColName} & &  \multicolumn{1}{c}{\textbf{\CPMIN}} & \multicolumn{1}{c}{\textbf{\CPMHI}} & & \multicolumn{1}{c}{\textbf{\CPMIN}} & \multicolumn{1}{c}{\textbf{\CPMHI}} \\
        \cmidrule{1-2} \cmidrule{4-5} \cmidrule{7-8}
        \multirow{3}{*}{\texttt{BERT}} 
        & \MetricL &  & 0.63 \ci{0.01} & 0.52 \ci{0.04} &  & 0.42 \ci{0.02} & 0.38  \ci{0.03} \\
        & \MetricCos &  & 0.46 \ci{0.00} & 0.45 \ci{0.01} &  & 0.34 \ci{0.02} & 0.30 \ci{0.06} \\
        & \MetricNormDiff &  & 0.39 \ci{0.01} & 0.33 \ci{0.02} &  & 0.23 \ci{0.01} & 0.22 \ci{0.05} \\
        \cmidrule{1-2} \cmidrule{4-5} \cmidrule{7-8}
        \multirow{3}{*}{\texttt{RoBERTa}} 
         & \MetricL &  & 0.66 \ci{0.01} & 0.63 \ci{0.04} &  & 0.40 \ci{0.01} & 0.37 \ci{0.04} \\
         & \MetricCos &  & 0.46 \ci{0.00} & 0.48 \ci{0.01} &  & 0.33 \ci{0.01} & 0.29 \ci{0.04} \\
         & \MetricNormDiff &  & 0.42 \ci{0.01} & 0.42 \ci{0.05} &  & 0.21 \ci{0.01} & 0.23 \ci{0.05} \\
        \cmidrule{1-2} \cmidrule{4-5} \cmidrule{7-8}
        \multirow{3}{*}{\texttt{GPT-2}} 
         & \MetricL &  & 0.55 \ci{0.01} & 0.41 \ci{0.03} &  & 0.38 \ci{0.01} & 0.36 \ci{0.04} \\
         & \MetricCos &  & 0.47 \ci{0.01} & 0.39 \ci{0.02} &  & 0.37 \ci{0.01} & 0.35 \ci{0.05} \\
         & \MetricNormDiff &  & 0.32 \ci{0.01} & 0.25 \ci{0.02} &  & 0.22 \ci{0.01} & 0.24 \ci{0.05} \\
        \cmidrule{1-2} \cmidrule{4-5} \cmidrule{7-8}
        \multirow{3}{*}{\texttt{LSTM}} 
         & \MetricL &  & 0.66 \ci{0.01} & 0.41 \ci{0.01} &  & 0.46 \ci{0.00} & 0.42 \ci{0.01} \\
         & \MetricCos &  & 0.50 \ci{0.00} & 0.42 \ci{0.02} &  & 0.50 \ci{0.02} & 0.40 \ci{0.01} \\
         & \MetricNormDiff &  & 0.42 \ci{0.00} & 0.25 \ci{0.00} &  & 0.31 \ci{0.00} & 0.28 \ci{0.02} \\
        \bottomrule
        \end{tabular}
    }
  \captionof{table}{Self-explanation CEBaB scores measured in three different metrics on the test set for four different model architectures as a five-class sentiment classification task. \textbf{Lower is better}. Results averaged over three distinct seeds, standard deviations in parentheses.}
  \label{tab:self-explain-baselines}
\end{minipage}
\end{minipage}
\end{figure}

\subsection{Concept-Aware Feature Attribution with \CPMHI{}}\label{sec:ig}

We have shown that \CPMHI{} provides trustworthy explanations (\secref{sec:result-cebab}). We now investigate whether \CPMHI{} learns representations that mediate the effects of different concepts. We adapt Integrated Gradients (IG;~\citealt{sundararajan17a}) to provide concept-aware feature attributions, by only considering gradients flowing through the hidden representation associated with a given concept. We formalize this version of IG in \appref{app:ig-method}.

In \tabref{tab:ig}, we compare concept-aware feature attibutions for two variants of \CPMHI{} (\texttt{IIT} and \texttt{Multi-task}) and the original black-box (\texttt{Finetuned}) model. For \texttt{IIT} we remove the multi-task objective $\mathcal{L}_{\text{Multi}}$ during training and for \texttt{Multi-task} we remove the the interchange intervention objective $\mathcal{L}_{\text{HI}}$. This helps isolate the individual effects of both losses on concept localization. All three models predict a neutral final sentiment score for the considered input, but they show vastly different feature attributions. Only \texttt{IIT} reliably highlights words that are semantically related to each concept. For instance, when we restrict the gradients to flow only through the intervention site of the \emph{noise} concept, ``loud'' is the word highlighted the most that contributes negatively. When we consider the \emph{service} concept, words like ``friendly'' and ``waiter'' are highlighted the most as contributing positively. 
These contrasts are missing for representations of the \texttt{Multi-task} and \texttt{Finetuned} models. Only the IIT training paradigm pushes the model to learn causally localized representations. For the \emph{service} concept, we notice that the \texttt{IIT} model wrongfully attributes ``delicious''. This could be useful for debugging purposes and could be used to highlight potential failure modes of the model.

\begin{table*}[tp]
\centering
\resizebox{0.99\linewidth}{!}{%
  \centering
  \setlength{\tabcolsep}{4pt}
  \begin{tabular}[c]{lllcc}
\toprule
Model & Predicted & Concept & Score & Word Importance \\
\midrule
\multirow{5}{*}{\texttt{Finetuned}} & \multirow{5}{*}{\emph{neutral}}
& ambiance & $+0.03$ & \colorbox{red!10}{\strut [CLS]}\colorbox{red!36}{\strut the}\colorbox{green!37}{\strut music}\colorbox{green!12}{\strut was}\colorbox{red!3}{\strut too}\colorbox{green!19}{\strut loud}\colorbox{red!20}{\strut ,}\colorbox{red!23}{\strut and}\colorbox{green!0}{\strut the}\colorbox{red!25}{\strut decorations}\colorbox{red!16}{\strut were}\colorbox{red!11}{\strut taste}\colorbox{green!41}{\strut \#\#less}\colorbox{red!32}{\strut ,}\colorbox{green!34}{\strut but}\colorbox{green!16}{\strut they}\colorbox{red!37}{\strut had}\colorbox{green!67}{\strut friendly}\colorbox{red!31}{\strut waiter}\colorbox{red!50}{\strut \#\#s}\colorbox{red!11}{\strut and}\colorbox{green!100}{\strut delicious}\colorbox{red!7}{\strut pasta}\colorbox{red!10}{\strut [SEP]} \\
& & food & $+0.11$ & \colorbox{red!14}{\strut [CLS]}\colorbox{green!14}{\strut the}\colorbox{red!38}{\strut music}\colorbox{red!43}{\strut was}\colorbox{green!62}{\strut too}\colorbox{green!33}{\strut loud}\colorbox{red!8}{\strut ,}\colorbox{red!18}{\strut and}\colorbox{green!35}{\strut the}\colorbox{red!5}{\strut decorations}\colorbox{red!11}{\strut were}\colorbox{red!11}{\strut taste}\colorbox{red!21}{\strut \#\#less}\colorbox{red!11}{\strut ,}\colorbox{red!3}{\strut but}\colorbox{green!12}{\strut they}\colorbox{red!37}{\strut had}\colorbox{green!100}{\strut friendly}\colorbox{red!51}{\strut waiter}\colorbox{green!10}{\strut \#\#s}\colorbox{green!2}{\strut and}\colorbox{green!68}{\strut delicious}\colorbox{red!39}{\strut pasta}\colorbox{red!24}{\strut [SEP]} \\
& & noise & $+0.04$ & \colorbox{green!6}{\strut [CLS]}\colorbox{green!24}{\strut the}\colorbox{red!19}{\strut music}\colorbox{red!3}{\strut was}\colorbox{green!11}{\strut too}\colorbox{green!13}{\strut loud}\colorbox{red!11}{\strut ,}\colorbox{red!4}{\strut and}\colorbox{green!5}{\strut the}\colorbox{red!6}{\strut decorations}\colorbox{red!10}{\strut were}\colorbox{red!12}{\strut taste}\colorbox{red!0}{\strut \#\#less}\colorbox{red!14}{\strut ,}\colorbox{green!4}{\strut but}\colorbox{green!12}{\strut they}\colorbox{red!51}{\strut had}\colorbox{green!100}{\strut friendly}\colorbox{red!23}{\strut waiter}\colorbox{red!34}{\strut \#\#s}\colorbox{red!58}{\strut and}\colorbox{green!72}{\strut delicious}\colorbox{green!5}{\strut pasta}\colorbox{red!3}{\strut [SEP]} \\
& & service & $+0.26$ & \colorbox{red!8}{\strut [CLS]}\colorbox{green!87}{\strut the}\colorbox{red!33}{\strut music}\colorbox{green!38}{\strut was}\colorbox{green!11}{\strut too}\colorbox{red!9}{\strut loud}\colorbox{red!54}{\strut ,}\colorbox{red!2}{\strut and}\colorbox{red!14}{\strut the}\colorbox{red!33}{\strut decorations}\colorbox{red!58}{\strut were}\colorbox{red!39}{\strut taste}\colorbox{green!49}{\strut \#\#less}\colorbox{red!14}{\strut ,}\colorbox{green!61}{\strut but}\colorbox{green!2}{\strut they}\colorbox{green!0}{\strut had}\colorbox{red!7}{\strut friendly}\colorbox{red!42}{\strut waiter}\colorbox{red!56}{\strut \#\#s}\colorbox{green!51}{\strut and}\colorbox{green!100}{\strut delicious}\colorbox{red!9}{\strut pasta}\colorbox{red!17}{\strut [SEP]} \\
\midrule
\multirow{5}{*}{\texttt{Multi-task}} & \multirow{5}{*}{\emph{neutral}}
& ambiance & $+0.25$ & \colorbox{red!13}{\strut [CLS]}\colorbox{red!11}{\strut the}\colorbox{red!12}{\strut music}\colorbox{red!14}{\strut was}\colorbox{green!61}{\strut too}\colorbox{green!42}{\strut loud}\colorbox{red!27}{\strut ,}\colorbox{red!6}{\strut and}\colorbox{red!3}{\strut the}\colorbox{red!12}{\strut decorations}\colorbox{red!4}{\strut were}\colorbox{red!36}{\strut taste}\colorbox{green!39}{\strut \#\#less}\colorbox{red!23}{\strut ,}\colorbox{red!0}{\strut but}\colorbox{red!23}{\strut they}\colorbox{red!33}{\strut had}\colorbox{green!41}{\strut friendly}\colorbox{red!2}{\strut waiter}\colorbox{red!17}{\strut \#\#s}\colorbox{red!3}{\strut and}\colorbox{green!100}{\strut delicious}\colorbox{red!22}{\strut pasta}\colorbox{red!13}{\strut [SEP]} \\
& & food & $+0.23$ & \colorbox{red!29}{\strut [CLS]}\colorbox{red!9}{\strut the}\colorbox{green!3}{\strut music}\colorbox{red!1}{\strut was}\colorbox{green!49}{\strut too}\colorbox{green!37}{\strut loud}\colorbox{red!25}{\strut ,}\colorbox{red!8}{\strut and}\colorbox{red!15}{\strut the}\colorbox{red!21}{\strut decorations}\colorbox{red!5}{\strut were}\colorbox{red!28}{\strut taste}\colorbox{green!21}{\strut \#\#less}\colorbox{red!13}{\strut ,}\colorbox{red!1}{\strut but}\colorbox{red!22}{\strut they}\colorbox{red!19}{\strut had}\colorbox{green!18}{\strut friendly}\colorbox{red!13}{\strut waiter}\colorbox{red!7}{\strut \#\#s}\colorbox{green!15}{\strut and}\colorbox{green!100}{\strut delicious}\colorbox{red!4}{\strut pasta}\colorbox{red!16}{\strut [SEP]} \\
& & noise & $+0.31$ & \colorbox{red!15}{\strut [CLS]}\colorbox{red!4}{\strut the}\colorbox{red!9}{\strut music}\colorbox{red!2}{\strut was}\colorbox{green!55}{\strut too}\colorbox{green!37}{\strut loud}\colorbox{red!28}{\strut ,}\colorbox{red!7}{\strut and}\colorbox{red!6}{\strut the}\colorbox{red!15}{\strut decorations}\colorbox{green!3}{\strut were}\colorbox{red!26}{\strut taste}\colorbox{green!25}{\strut \#\#less}\colorbox{red!19}{\strut ,}\colorbox{red!4}{\strut but}\colorbox{red!20}{\strut they}\colorbox{red!31}{\strut had}\colorbox{green!11}{\strut friendly}\colorbox{red!13}{\strut waiter}\colorbox{red!10}{\strut \#\#s}\colorbox{red!6}{\strut and}\colorbox{green!100}{\strut delicious}\colorbox{green!7}{\strut pasta}\colorbox{red!16}{\strut [SEP]} \\
& & service & $+0.16$ & \colorbox{red!17}{\strut [CLS]}\colorbox{green!5}{\strut the}\colorbox{green!1}{\strut music}\colorbox{red!9}{\strut was}\colorbox{green!52}{\strut too}\colorbox{green!32}{\strut loud}\colorbox{red!23}{\strut ,}\colorbox{red!4}{\strut and}\colorbox{red!7}{\strut the}\colorbox{red!7}{\strut decorations}\colorbox{red!9}{\strut were}\colorbox{red!21}{\strut taste}\colorbox{green!29}{\strut \#\#less}\colorbox{red!21}{\strut ,}\colorbox{green!11}{\strut but}\colorbox{red!19}{\strut they}\colorbox{red!27}{\strut had}\colorbox{green!8}{\strut friendly}\colorbox{red!16}{\strut waiter}\colorbox{red!20}{\strut \#\#s}\colorbox{red!2}{\strut and}\colorbox{green!100}{\strut delicious}\colorbox{red!18}{\strut pasta}\colorbox{red!11}{\strut [SEP]} \\
\midrule
\multirow{5}{*}{\texttt{IIT}} & \multirow{5}{*}{\emph{neutral}}
& ambiance & $-0.24$ & \colorbox{green!0}{\strut [CLS]}\colorbox{red!44}{\strut the}\colorbox{green!55}{\strut music}\colorbox{red!71}{\strut was}\colorbox{green!100}{\strut too}\colorbox{red!9}{\strut loud}\colorbox{red!6}{\strut ,}\colorbox{red!25}{\strut and}\colorbox{red!6}{\strut the}\colorbox{red!44}{\strut decorations}\colorbox{red!22}{\strut were}\colorbox{green!41}{\strut taste}\colorbox{green!37}{\strut \#\#less}\colorbox{red!0}{\strut ,}\colorbox{red!12}{\strut but}\colorbox{green!30}{\strut they}\colorbox{green!37}{\strut had}\colorbox{red!16}{\strut friendly}\colorbox{red!95}{\strut waiter}\colorbox{green!5}{\strut \#\#s}\colorbox{green!18}{\strut and}\colorbox{green!18}{\strut delicious}\colorbox{green!8}{\strut pasta}\colorbox{green!0}{\strut [SEP]} \\
& & food & $+1.11$ & \colorbox{red!8}{\strut [CLS]}\colorbox{red!2}{\strut the}\colorbox{green!0}{\strut music}\colorbox{green!4}{\strut was}\colorbox{green!1}{\strut too}\colorbox{green!7}{\strut loud}\colorbox{red!13}{\strut ,}\colorbox{red!7}{\strut and}\colorbox{red!14}{\strut the}\colorbox{green!19}{\strut decorations}\colorbox{red!3}{\strut were}\colorbox{red!22}{\strut taste}\colorbox{red!13}{\strut \#\#less}\colorbox{red!9}{\strut ,}\colorbox{green!5}{\strut but}\colorbox{red!9}{\strut they}\colorbox{red!12}{\strut had}\colorbox{green!12}{\strut friendly}\colorbox{red!29}{\strut waiter}\colorbox{red!10}{\strut \#\#s}\colorbox{green!4}{\strut and}\colorbox{green!100}{\strut delicious}\colorbox{green!7}{\strut pasta}\colorbox{red!5}{\strut [SEP]} \\
& & noise & $-0.98$ & \colorbox{green!32}{\strut [CLS]}\colorbox{green!10}{\strut the}\colorbox{red!38}{\strut music}\colorbox{red!15}{\strut was}\colorbox{red!100}{\strut too}\colorbox{red!84}{\strut loud}\colorbox{green!0}{\strut ,}\colorbox{green!11}{\strut and}\colorbox{green!1}{\strut the}\colorbox{green!32}{\strut decorations}\colorbox{green!5}{\strut were}\colorbox{green!28}{\strut taste}\colorbox{green!24}{\strut \#\#less}\colorbox{green!10}{\strut ,}\colorbox{green!13}{\strut but}\colorbox{green!41}{\strut they}\colorbox{green!9}{\strut had}\colorbox{red!16}{\strut friendly}\colorbox{green!16}{\strut waiter}\colorbox{green!14}{\strut \#\#s}\colorbox{green!16}{\strut and}\colorbox{red!21}{\strut delicious}\colorbox{red!17}{\strut pasta}\colorbox{green!24}{\strut [SEP]} \\
& & service & $+1.16$ & \colorbox{red!12}{\strut [CLS]}\colorbox{red!5}{\strut the}\colorbox{red!37}{\strut music}\colorbox{red!10}{\strut was}\colorbox{green!7}{\strut too}\colorbox{green!5}{\strut loud}\colorbox{green!19}{\strut ,}\colorbox{red!18}{\strut and}\colorbox{green!6}{\strut the}\colorbox{red!8}{\strut decorations}\colorbox{red!33}{\strut were}\colorbox{red!13}{\strut taste}\colorbox{red!2}{\strut \#\#less}\colorbox{red!10}{\strut ,}\colorbox{red!8}{\strut but}\colorbox{red!7}{\strut they}\colorbox{red!1}{\strut had}\colorbox{green!78}{\strut friendly}\colorbox{green!100}{\strut waiter}\colorbox{green!33}{\strut \#\#s}\colorbox{red!0}{\strut and}\colorbox{red!60}{\strut delicious}\colorbox{red!5}{\strut pasta}\colorbox{red!13}{\strut [SEP]} \\
\bottomrule
\end{tabular}}
  \caption{Visualizations of word importance scores using Integrated Gradient (IG) by restricting gradient flow through the corresponding intervention site of the targeted concept. 
  Our target class pools \emph{positive} and  \emph{very positive}. Individual word importance is the sum of neuron-level importance scores for each input, normalized to [$-1$, $+1$]. $-1$ means the word contributes the most negatively to predicting the target class (red); $+1$ means the word contributes the most positively %
  (green).}
  \label{tab:ig}
\end{table*}

\section{Conclusion}\label{sec:conclusion}
We explored the use of approximate counterfactual training data to build more robust causal explanation methods. We introduced \OurMethod s (\OurMethodAbbr s), which learn to mimic both the \textit{factual} and \textit{counterfactual} behaviors of a black-box model $\bbm{}$. Using CEBaB, a benchmark for causal concept-based explanation methods, we demonstrated that both versions of our technique (\CPMIN{} and \CPMHI{}) significantly outperform previous explanation methods. 

Interestingly, we find that our \texttt{GPT-3} based explanation method performs on-par with our best \OurMethodAbbr{} model in some settings. While test-time use of \texttt{GPT-3} as explanation method might not be feasible, we believe this result shows that \texttt{GPT-3} could be deployed to supplement human-annotation efforts for counterfactual data creation. 

Our results suggest that \OurMethodAbbr s can be more than just explanation methods. They achieve factual performance on par with the model they aim to explain, and they can explain their own behavior. This paves the way to using them as deployed models that both perform tasks and offer explanations. 
In addition, the causally localized representations of our \CPMHI{} variant are very intuitive, as revealed by our concept-aware feature attribution technique. We believe that causal localization techniques could play a vital role in further model explanation efforts.

\section{Acknowledgement}
This research is supported in part by a grant from Meta AI. Karel D’Oosterlinck was supported through a doctoral fellowship from the Special Research Fund (BOF) of Ghent University.

\bibliography{iclr2023_conference, custom, anthology}

\begin{thebibliography}{56}
\providecommand{\natexlab}[1]{#1}
\providecommand{\url}[1]{\texttt{#1}}
\expandafter\ifx\csname urlstyle\endcsname\relax
  \providecommand{\doi}[1]{doi: #1}\else
  \providecommand{\doi}{doi: \begingroup \urlstyle{rm}\Url}\fi

\bibitem[Abraham et~al.(2022)Abraham, D'Oosterlinck, Feder, Gat, Geiger, Potts,
  Reichart, and Wu]{abraham2022cebab}
Eldar~David Abraham, Karel D'Oosterlinck, Amir Feder, Yair~Ori Gat, Atticus
  Geiger, Christopher Potts, Roi Reichart, and Zhengxuan Wu.
\newblock {CEBaB}: Estimating the causal effects of real-world concepts on
  {NLP} model behavior.
\newblock \emph{Advances in Neural Information Processing Systems}, 2022.
\newblock URL \url{https://arxiv.org/abs/2205.14140}.

\bibitem[Amodei et~al.(2016)Amodei, Olah, Steinhardt, Christiano, Schulman, and
  Mané]{amodei_concrete_2016}
Dario Amodei, Chris Olah, Jacob Steinhardt, Paul Christiano, John Schulman, and
  Dan Mané.
\newblock Concrete problems in {AI} safety.
\newblock \emph{{}}, 2016.
\newblock URL \url{http://arxiv.org/abs/1606.06565}.

\bibitem[Ban et~al.(2022)Ban, Jiang, Liu, and
  Steinert-Threlkeld]{ban2022testing}
Pangbo Ban, Yifan Jiang, Tianran Liu, and Shane Steinert-Threlkeld.
\newblock Testing pre-trained language models' understanding of distributivity
  via causal mediation analysis.
\newblock arXiv:2209.04761, 2022.
\newblock URL \url{https://arxiv.org/abs/2209.04761}.

\bibitem[Binder et~al.(2016)Binder, Montavon, Lapuschkin, M{\"u}ller, and
  Samek]{Binder16}
Alexander Binder, Gr{\'e}goire Montavon, Sebastian Lapuschkin, Klaus-Robert
  M{\"u}ller, and Wojciech Samek.
\newblock Layer-wise relevance propagation for neural networks with local
  renormalization layers.
\newblock In \emph{International Conference on Artificial Neural Networks},
  2016.
\newblock URL \url{https://doi.org/10.1007/978-3-319-44781-0_8}.

\bibitem[Brown et~al.(2020)Brown, Mann, Ryder, Subbiah, Kaplan, Dhariwal,
  Neelakantan, Shyam, Sastry, Askell, et~al.]{brown_language_2020}
Tom Brown, Benjamin Mann, Nick Ryder, Melanie Subbiah, Jared~D Kaplan, Prafulla
  Dhariwal, Arvind Neelakantan, Pranav Shyam, Girish Sastry, Amanda Askell,
  et~al.
\newblock Language models are few-shot learners.
\newblock \emph{Advances in Neural Information Processing Systems}, 2020.
\newblock URL
  \url{https://proceedings.neurips.cc/paper/2020/file/1457c0d6bfcb4967418bfb8ac142f64a-Paper.pdf}.

\bibitem[Buitinck et~al.(2013)Buitinck, Louppe, Blondel, Pedregosa, Mueller,
  Grisel, Niculae, Prettenhofer, Gramfort, Grobler, Layton, VanderPlas, Joly,
  Holt, and Varoquaux]{sklearn_api}
Lars Buitinck, Gilles Louppe, Mathieu Blondel, Fabian Pedregosa, Andreas
  Mueller, Olivier Grisel, Vlad Niculae, Peter Prettenhofer, Alexandre
  Gramfort, Jaques Grobler, Robert Layton, Jake VanderPlas, Arnaud Joly, Brian
  Holt, and Ga{\"{e}}l Varoquaux.
\newblock {API} design for machine learning software: Experiences from the
  scikit-learn project.
\newblock In \emph{ECML PKDD Workshop: Languages for Data Mining and Machine
  Learning}, 2013.
\newblock URL \url{https://hal.inria.fr/hal-00856511}.

\bibitem[Clark et~al.(2019)Clark, Khandelwal, Levy, and
  Manning]{clark-etal-2019-bert}
Kevin Clark, Urvashi Khandelwal, Omer Levy, and Christopher~D. Manning.
\newblock What does {BERT} look at? {An} analysis of {BERT}{'}s attention.
\newblock In \emph{Proceedings of the 2019 ACL Workshop BlackboxNLP: Analyzing
  and Interpreting Neural Networks for NLP}, Florence, Italy, 2019.
\newblock URL \url{https://www.aclweb.org/anthology/W19-4828}.

\bibitem[Conneau et~al.(2018)Conneau, Kruszewski, Lample, Barrault, and
  Baroni]{conneau-etal-2018-cram}
Alexis Conneau, German Kruszewski, Guillaume Lample, Lo{\"\i}c Barrault, and
  Marco Baroni.
\newblock What you can cram into a single {\$}{\&}!{\#}* vector: Probing
  sentence embeddings for linguistic properties.
\newblock In \emph{Proceedings of the 56th Annual Meeting of the Association
  for Computational Linguistics}, Melbourne, Australia, 2018.
\newblock URL \url{https://www.aclweb.org/anthology/P18-1198}.

\bibitem[De~Cao et~al.(2021)De~Cao, Schmid, Hupkes, and Titov]{de2021sparse}
Nicola De~Cao, Leon Schmid, Dieuwke Hupkes, and Ivan Titov.
\newblock Sparse interventions in language models with differentiable masking.
\newblock arxiv:2112.06837, 2021.
\newblock URL \url{https://arxiv.org/abs/2112.06837}.

\bibitem[Devlin et~al.(2019)Devlin, Chang, Lee, and
  Toutanova]{devlin_bert_2019}
Jacob Devlin, Ming-Wei Chang, Kenton Lee, and Kristina Toutanova.
\newblock {BERT}: Pre-training of deep bidirectional transformers for language
  understanding.
\newblock In \emph{Proceedings of the 2019 Conference of the North {A}merican
  Chapter of the Association for Computational Linguistics: Human Language
  Technologies}, Minneapolis, Minnesota, 2019.
\newblock URL \url{https://www.aclweb.org/anthology/N19-1423}.

\bibitem[Ehsan et~al.(2021)Ehsan, Liao, Muller, Riedl, and
  Weisz]{ehsan2021expanding}
Upol Ehsan, Q~Vera Liao, Michael Muller, Mark~O Riedl, and Justin~D Weisz.
\newblock Expanding explainability: Towards social transparency in {AI}
  systems.
\newblock In \emph{Proceedings of the 2021 CHI Conference on Human Factors in
  Computing Systems}, 2021.
\newblock URL \url{https://dl.acm.org/doi/pdf/10.1145/3411764.3445188}.

\bibitem[Elazar et~al.(2021)Elazar, Ravfogel, Jacovi, and
  Goldberg]{elazar2021amnesic}
Yanai Elazar, Shauli Ravfogel, Alon Jacovi, and Yoav Goldberg.
\newblock Amnesic probing: Behavioral explanation with amnesic counterfactuals.
\newblock \emph{Transactions of the Association for Computational Linguistics},
  2021.
\newblock URL
  \url{https://direct.mit.edu/tacl/article/doi/10.1162/tacl_a_00359/98091/Amnesic-Probing-Behavioral-Explanation-with}.

\bibitem[Feder et~al.(2020)Feder, Oved, Shalit, and
  Reichart]{feder_causalm_2020}
Amir Feder, Nadav Oved, Uri Shalit, and Roi Reichart.
\newblock {C}ausa{LM}: Causal model explanation through counterfactual language
  models.
\newblock \emph{Computational Linguistics}, 2020.
\newblock URL \url{https://aclanthology.org/2021.cl-2.13}.

\bibitem[Geiger et~al.(2020)Geiger, Richardson, and
  Potts]{geiger-etal-2020-neural}
Atticus Geiger, Kyle Richardson, and Christopher Potts.
\newblock Neural natural language inference models partially embed theories of
  lexical entailment and negation.
\newblock In \emph{Proceedings of the Third BlackboxNLP Workshop on Analyzing
  and Interpreting Neural Networks for NLP}, 2020.
\newblock URL \url{https://aclanthology.org/2020.blackboxnlp-1.16}.

\bibitem[Geiger et~al.(2021)Geiger, Lu, Icard, and Potts]{geiger2021causal}
Atticus Geiger, Hanson Lu, Thomas Icard, and Christopher Potts.
\newblock Causal abstractions of neural networks.
\newblock \emph{Advances in Neural Information Processing Systems}, 2021.
\newblock URL
  \url{https://proceedings.neurips.cc/paper/2021/file/4f5c422f4d49a5a807eda27434231040-Paper.pdf}.

\bibitem[Geiger et~al.(2022)Geiger, Wu, Lu, Rozner, Kreiss, Icard, Goodman, and
  Potts]{geiger-etal-2021-iit}
Atticus Geiger, Zhengxuan Wu, Hanson Lu, Josh Rozner, Elisa Kreiss, Thomas
  Icard, Noah Goodman, and Christopher Potts.
\newblock Inducing causal structure for interpretable neural networks.
\newblock In \emph{International Conference on Machine Learning}, 2022.
\newblock URL \url{https://proceedings.mlr.press/v162/geiger22a.html}.

\bibitem[Goodman \& Flaxman(2017)Goodman and Flaxman]{goodman_european_2017}
Bryce Goodman and Seth Flaxman.
\newblock European {Union} regulations on algorithmic decision-making and a
  ``right to explanation''.
\newblock \emph{AI Magazine}, 2017.
\newblock URL \url{http://arxiv.org/abs/1606.08813}.

\bibitem[Goyal et~al.(2019)Goyal, Wu, Ernst, Batra, Parikh, and
  Lee]{goyal2019counterfactual}
Yash Goyal, Ziyan Wu, Jan Ernst, Dhruv Batra, Devi Parikh, and Stefan Lee.
\newblock Counterfactual visual explanations.
\newblock In \emph{International Conference on Machine Learning}, 2019.
\newblock URL \url{http://proceedings.mlr.press/v97/goyal19a.html}.

\bibitem[Guidotti et~al.(2018)Guidotti, Monreale, Ruggieri, Turini, Giannotti,
  and Pedreschi]{guidotti2018survey}
Riccardo Guidotti, Anna Monreale, Salvatore Ruggieri, Franco Turini, Fosca
  Giannotti, and Dino Pedreschi.
\newblock A survey of methods for explaining black box models.
\newblock \emph{ACM computing surveys (CSUR)}, 2018.
\newblock URL \url{https://dl.acm.org/doi/abs/10.1145/3236009}.

\bibitem[Hardt et~al.(2016)Hardt, Price, and Srebro]{hardt_equality_2016}
Moritz Hardt, Eric Price, and Nati Srebro.
\newblock Equality of opportunity in supervised learning.
\newblock In \emph{Advances in {Neural} {Information} {Processing} {Systems}}.
  Curran Associates, Inc., 2016.
\newblock URL
  \url{https://proceedings.neurips.cc/paper/2016/file/9d2682367c3935defcb1f9e247a97c0d-Paper.pdf}.

\bibitem[Hinton et~al.(2015)Hinton, Vinyals, Dean,
  et~al.]{hinton2015distilling}
Geoffrey Hinton, Oriol Vinyals, Jeff Dean, et~al.
\newblock Distilling the knowledge in a neural network.
\newblock \emph{NeurIPS Deep Learning and Representation Learning Workshop},
  2015.
\newblock URL \url{https://arxiv.org/abs/1503.02531}.

\bibitem[Hochreiter \& Schmidhuber(1997)Hochreiter and
  Schmidhuber]{Hochreiter:Schmidhuber:1997}
Sepp Hochreiter and J{\"u}rgen Schmidhuber.
\newblock Long short-term memory.
\newblock \emph{Neural Computation}, 1997.
\newblock URL \url{https://ieeexplore.ieee.org/abstract/document/6795963}.

\bibitem[Holland(1986)]{holland1986statistics}
Paul~W Holland.
\newblock Statistics and causal inference.
\newblock \emph{Journal of the American statistical Association}, 1986.
\newblock URL
  \url{https://www.tandfonline.com/doi/abs/10.1080/01621459.1986.10478354}.

\bibitem[Jacovi \& Goldberg(2020)Jacovi and Goldberg]{jacovi2020towards}
Alon Jacovi and Yoav Goldberg.
\newblock Towards faithfully interpretable {NLP} systems: How should we define
  and evaluate faithfulness?
\newblock In \emph{Proceedings of the 58th Annual Meeting of the Association
  for Computational Linguistics}, 2020.
\newblock URL \url{https://aclanthology.org/2020.acl-main.386}.

\bibitem[Jakesch et~al.(2019)Jakesch, French, Ma, Hancock, and
  Naaman]{jakesch2019ai}
Maurice Jakesch, Megan French, Xiao Ma, Jeffrey~T Hancock, and Mor Naaman.
\newblock {AI}-mediated communication: How the perception that profile text was
  written by {AI} affects trustworthiness.
\newblock In \emph{Proceedings of the 2019 CHI Conference on Human Factors in
  Computing Systems}, 2019.
\newblock URL \url{https://dl.acm.org/doi/abs/10.1145/3290605.3300469}.

\bibitem[Kim(2015)]{kim2015}
Been Kim.
\newblock \emph{Interactive and Interpretable Machine Learning Models for Human
  Machine Collaboration}.
\newblock PhD thesis, Massachusetts Institute of Technology, 2015.
\newblock URL \url{https://dspace.mit.edu/handle/1721.1/98680}.

\bibitem[Kim et~al.(2018)Kim, Wattenberg, Gilmer, Cai, Wexler, Viegas, and
  Sayres]{kim_interpretability_2018}
Been Kim, Martin Wattenberg, Justin Gilmer, Carrie Cai, James Wexler, Fernanda
  Viegas, and Rory Sayres.
\newblock Interpretability beyond feature attribution: Quantitative testing
  with concept activation vectors ({TCAV}).
\newblock In \emph{International {Conference} on {Machine} {Learning}}, 2018.
\newblock URL \url{http://proceedings.mlr.press/v80/kim18d.html}.

\bibitem[Kleinberg et~al.(2017)Kleinberg, Mullainathan, and
  Raghavan]{kleinberg2017}
Jon Kleinberg, Sendhil Mullainathan, and Manish Raghavan.
\newblock Inherent trade-offs in the fair determination of risk scores.
\newblock In \emph{8th Innovations in Theoretical Computer Science Conference},
  2017.
\newblock URL \url{http://drops.dagstuhl.de/opus/volltexte/2017/8156}.

\bibitem[Koh et~al.(2020)Koh, Nguyen, Tang, Mussmann, Pierson, Kim, and
  Liang]{koh2020concept}
Pang~Wei Koh, Thao Nguyen, Yew~Siang Tang, Stephen Mussmann, Emma Pierson, Been
  Kim, and Percy Liang.
\newblock Concept bottleneck models.
\newblock In \emph{International Conference on Machine Learning}, 2020.
\newblock URL \url{https://proceedings.mlr.press/v119/koh20a.html}.

\bibitem[K{\"u}nzel et~al.(2019)K{\"u}nzel, Sekhon, Bickel, and
  Yu]{kunzel2019metalearners}
S{\"o}ren~R K{\"u}nzel, Jasjeet~S Sekhon, Peter~J Bickel, and Bin Yu.
\newblock Metalearners for estimating heterogeneous treatment effects using
  machine learning.
\newblock \emph{Proceedings of the National Academy of Sciences}, 2019.

\bibitem[Lipton(2018)]{lipton_mythos_2018}
Zachary~C. Lipton.
\newblock The mythos of model interpretability.
\newblock \emph{Communications of the ACM}, 2018.
\newblock URL \url{https://doi.org/10.1145/3233231}.

\bibitem[Liu et~al.(2019)Liu, Ott, Goyal, Du, Joshi, Chen, Levy, Lewis,
  Zettlemoyer, and Stoyanov]{liu_roberta_2019}
Yinhan Liu, Myle Ott, Naman Goyal, Jingfei Du, Mandar Joshi, Danqi Chen, Omer
  Levy, Mike Lewis, Luke Zettlemoyer, and Veselin Stoyanov.
\newblock {RoBERTa}: {A} robustly optimized {BERT} pretraining approach.
\newblock \emph{{}}, 2019.
\newblock URL \url{http://arxiv.org/abs/1907.11692}.

\bibitem[Lovering \& Pavlick(2022)Lovering and Pavlick]{lovering2022unit}
Charles Lovering and Ellie Pavlick.
\newblock Unit testing for concepts in neural networks.
\newblock \emph{arXiv preprint arXiv:2208.10244}, 2022.

\bibitem[Manning et~al.(2020)Manning, Clark, Hewitt, Khandelwal, and
  Levy]{Manning-etal:2020}
Christopher~D. Manning, Kevin Clark, John Hewitt, Urvashi Khandelwal, and Omer
  Levy.
\newblock Emergent linguistic structure in artificial neural networks trained
  by self-supervision.
\newblock \emph{Proceedings of the National Academy of Sciences}, 2020.
\newblock URL \url{https://www.pnas.org/content/117/48/30046}.

\bibitem[Mehrabi et~al.(2021)Mehrabi, Morstatter, Saxena, Lerman, and
  Galstyan]{mehrabi2021survey}
Ninareh Mehrabi, Fred Morstatter, Nripsuta Saxena, Kristina Lerman, and Aram
  Galstyan.
\newblock A survey on bias and fairness in machine learning.
\newblock \emph{ACM Computing Surveys (CSUR)}, 2021.
\newblock URL \url{https://dl.acm.org/doi/abs/10.1145/3457607}.

\bibitem[Molnar(2020)]{molnar2020interpretable}
Christoph Molnar.
\newblock \emph{Interpretable Machine Learning}.
\newblock {}, 2020.

\bibitem[Otte(2013)]{otte_safe_2013}
Clemens Otte.
\newblock Safe and interpretable machine learning: A methodological review.
\newblock In \emph{Computational {Intelligence} in {Intelligent} {Data}
  {Analysis}}, 2013.
\newblock URL
  \url{https://link.springer.com/chapter/10.1007/978-3-642-32378-2_8}.

\bibitem[Paszke et~al.(2019)Paszke, Gross, Massa, Lerer, Bradbury, Chanan,
  Killeen, Lin, Gimelshein, Antiga, et~al.]{paszke2019pytorch}
Adam Paszke, Sam Gross, Francisco Massa, Adam Lerer, James Bradbury, Gregory
  Chanan, Trevor Killeen, Zeming Lin, Natalia Gimelshein, Luca Antiga, et~al.
\newblock Pytorch: An imperative style, high-performance deep learning library.
\newblock \emph{Advances in Neural Information Processing Systems}, 2019.

\bibitem[Pearl(2019)]{pearl_limitations_2019}
Judea Pearl.
\newblock The limitations of opaque learning machines.
\newblock \emph{{Possible Minds: Twenty-Five Ways of Looking at AI}}, 2019.
\newblock URL \url{https://ftp.cs.ucla.edu/pub/stat_ser/r489.pdf}.

\bibitem[Pennington et~al.(2014)Pennington, Socher, and
  Manning]{pennington-etal-2014-glove}
Jeffrey Pennington, Richard Socher, and Christopher Manning.
\newblock {G}lo{V}e: Global vectors for word representation.
\newblock In \emph{Proceedings of the 2014 Conference on Empirical Methods in
  Natural Language Processing ({EMNLP})}, 2014.
\newblock URL \url{https://aclanthology.org/D14-1162}.

\bibitem[Radford et~al.(2019)Radford, Wu, Child, Luan, Amodei, and
  Sutskever]{radford2019language}
Alec Radford, Jeffrey Wu, Rewon Child, David Luan, Dario Amodei, and Ilya
  Sutskever.
\newblock Language models are unsupervised multitask learners.
\newblock \emph{OpenAI blog}, 2019.
\newblock URL
  \url{https://cdn.openai.com/better-language-models/language_models_are_unsupervised_multitask_learners.pdf}.

\bibitem[Ravfogel et~al.(2020)Ravfogel, Elazar, Gonen, Twiton, and
  Goldberg]{ravfogel_null_2020}
Shauli Ravfogel, Yanai Elazar, Hila Gonen, Michael Twiton, and Yoav Goldberg.
\newblock Null it out: Guarding protected attributes by iterative nullspace
  projection.
\newblock In \emph{Proceedings of the 58th Annual Meeting of the Association
  for Computational Linguistics}, 2020.
\newblock URL \url{https://aclanthology.org/2020.acl-main.647}.

\bibitem[Ribeiro et~al.(2016)Ribeiro, Singh, and Guestrin]{ribeiro_why_2016}
Marco~Tulio Ribeiro, Sameer Singh, and Carlos Guestrin.
\newblock "{Why} {Should} {I} {Trust} {You}?": Explaining the predictions of
  any classifier.
\newblock In \emph{Proceedings of the 22nd {ACM} {SIGKDD} {International}
  {Conference} on {Knowledge} {Discovery} and {Data} {Mining}}, San Francisco
  California USA, 2016.
\newblock URL \url{https://dl.acm.org/doi/10.1145/2939672.2939778}.

\bibitem[Saphra \& Lopez(2019)Saphra and
  Lopez]{saphra-lopez-2019-understanding}
Naomi Saphra and Adam Lopez.
\newblock Understanding learning dynamics of language models with {SVCCA}.
\newblock In \emph{Proceedings of the 2019 Conference of the North {A}merican
  Chapter of the Association for Computational Linguistics: Human Language
  Technologies}, 2019.
\newblock URL \url{https://www.aclweb.org/anthology/N19-1329}.

\bibitem[Shrikumar et~al.(2017)Shrikumar, Greenside, and Kundaje]{Shrikumar16}
Avanti Shrikumar, Peyton Greenside, and Anshul Kundaje.
\newblock {Learning Important Features through Propagating Activation
  Differences}.
\newblock In \emph{International Conference on Machine Learning}, 2017.
\newblock URL \url{http://proceedings.mlr.press/v70/shrikumar17a}.

\bibitem[Soulos et~al.(2020)Soulos, McCoy, Linzen, and
  Smolensky]{soulos-etal-2020-discovering}
Paul Soulos, R.~Thomas McCoy, Tal Linzen, and Paul Smolensky.
\newblock Discovering the compositional structure of vector representations
  with role learning networks.
\newblock In \emph{Proceedings of the Third BlackboxNLP Workshop on Analyzing
  and Interpreting Neural Networks for NLP}, Online, 2020.
\newblock URL \url{https://aclanthology.org/2020.blackboxnlp-1.23}.

\bibitem[Springenberg et~al.(2014)Springenberg, Dosovitskiy, Brox, and
  Riedmiller]{springerberg2014}
Jost Springenberg, Alexey Dosovitskiy, Thomas Brox, and Martin Riedmiller.
\newblock Striving for simplicity: the all convolutional net.
\newblock \emph{CoRR}, 12 2014.
\newblock URL \url{https://arxiv.org/abs/1412.6806}.

\bibitem[Sundararajan et~al.(2017)Sundararajan, Taly, and Yan]{sundararajan17a}
Mukund Sundararajan, Ankur Taly, and Qiqi Yan.
\newblock Axiomatic attribution for deep networks.
\newblock In \emph{Proceedings of the 34th International Conference on Machine
  Learning - Volume 70}, 2017.
\newblock URL \url{http://proceedings.mlr.press/v70/sundararajan17a.html}.

\bibitem[Tenney et~al.(2019)Tenney, Das, and Pavlick]{tenney-etal-2019-bert}
Ian Tenney, Dipanjan Das, and Ellie Pavlick.
\newblock {BERT} rediscovers the classical {NLP} pipeline.
\newblock In \emph{Proceedings of the 57th Annual Meeting of the Association
  for Computational Linguistics}, 2019.
\newblock URL \url{https://www.aclweb.org/anthology/P19-1452}.

\bibitem[Vaswani et~al.(2017)Vaswani, Shazeer, Parmar, Uszkoreit, Jones, Gomez,
  Kaiser, and Polosukhin]{Vaswani-etal:2017}
Ashish Vaswani, Noam Shazeer, Niki Parmar, Jakob Uszkoreit, Llion Jones,
  Aidan~N Gomez, \L{}ukasz Kaiser, and Illia Polosukhin.
\newblock Attention is all you need.
\newblock In \emph{Advances in Neural Information Processing Systems 30}. {},
  2017.
\newblock URL
  \url{http://papers.nips.cc/paper/7181-attention-is-all-you-need.pdf}.

\bibitem[Verma et~al.(2020)Verma, Dickerson, and
  Hines]{verma2020counterfactual}
Sahil Verma, John Dickerson, and Keegan Hines.
\newblock Counterfactual explanations for machine learning: A review.
\newblock \emph{{}}, 2020.
\newblock URL \url{https://arxiv.org/abs/2010.10596}.

\bibitem[Vig et~al.(2020)Vig, Gehrmann, Belinkov, Qian, Nevo, Singer, and
  Shieber]{vig2020causal}
Jesse Vig, Sebastian Gehrmann, Yonatan Belinkov, Sharon Qian, Daniel Nevo,
  Yaron Singer, and Stuart~M. Shieber.
\newblock Investigating gender bias in language models using causal mediation
  analysis.
\newblock In \emph{Advances in Neural Information Processing Systems}, 2020.
\newblock URL
  \url{https://proceedings.neurips.cc/paper/2020/hash/92650b2e92217715fe312e6fa7b90d82-Abstract.html}.

\bibitem[Wolf et~al.(2019)Wolf, Debut, Sanh, Chaumond, Delangue, Moi, Cistac,
  Rault, Louf, Funtowicz, et~al.]{wolf2019huggingface}
Thomas Wolf, Lysandre Debut, Victor Sanh, Julien Chaumond, Clement Delangue,
  Anthony Moi, Pierric Cistac, Tim Rault, R{\'e}mi Louf, Morgan Funtowicz,
  et~al.
\newblock Huggingface's transformers: State-of-the-art natural language
  processing.
\newblock \emph{{}}, 2019.
\newblock URL \url{https://arxiv.org/abs/1910.03771}.

\bibitem[Wu et~al.(2021)Wu, Ribeiro, Heer, and Weld]{wu2021polyjuice}
Tongshuang Wu, Marco~Tulio Ribeiro, Jeffrey Heer, and Daniel Weld.
\newblock Polyjuice: Generating counterfactuals for explaining, evaluating, and
  improving models.
\newblock In \emph{Proceedings of the 59th Annual Meeting of the Association
  for Computational Linguistics and the 11th International Joint Conference on
  Natural Language Processing}, 2021.

\bibitem[Yeh et~al.(2020)Yeh, Kim, Arik, Li, Pfister, and
  Ravikumar]{yeh2020completeness}
Chih-Kuan Yeh, Been Kim, Sercan Arik, Chun-Liang Li, Tomas Pfister, and Pradeep
  Ravikumar.
\newblock On completeness-aware concept-based explanations in deep neural
  networks.
\newblock \emph{Advances in Neural Information Processing Systems}, 2020.
\newblock URL
  \url{https://proceedings.neurips.cc/paper/2020/file/ecb287ff763c169694f682af52c1f309-Paper.pdf}.

\bibitem[Zeiler \& Fergus(2014)Zeiler and Fergus]{Zeiler2014}
Matthew~D. Zeiler and Rob Fergus.
\newblock Visualizing and understanding convolutional networks.
\newblock In David Fleet, Tomas Pajdla, Bernt Schiele, and Tinne Tuytelaars
  (eds.), \emph{European Conference on Computer Vision}, 2014.
\newblock URL
  \url{https://link.springer.com/chapter/10.1007/978-3-319-10590-1_53}.

\end{thebibliography}
\bibliographystyle{iclr2023_conference}

\appendix
\section{Appendix}

\subsection{CEBaB Dataset Statistics} \label{app:dataset-stats}

\Tabref{tab:cebab-stats} shows dataset statistics of CEBaB. The variants of CEBaB we consider only impact the train split. The top panel shows the number of observational samples and edits introduced in the CEBaB paper. The bottom panel shows our \textit{paired} versions, where we create approximate counterfactual pairs. We explore two variants of approximate counterfactuals: \emph{human}-created and \emph{sampled} counterfactuals (\secref{subsec:cebab}). The \emph{human} setting considers all pairs made possible by using \textit{all} data. The \emph{sampling} setting considers pairs sampled from only the \emph{observational} data, as discussed in \secref{app:pairs-sampling}.

\begin{table*}[ht]
\centering
\resizebox{0.55\linewidth}{!}{%
  \centering
  \setlength{\tabcolsep}{4pt}
  \begin{tabular}[c]{l c c c c}
    \toprule
        \textbf{Dataset} & & \# train & \# dev & \# test  \\
    \cmidrule{1-1} \cmidrule{3-5}
        CEBaB (\emph{observational}) & & 1,755 & 1,673 & 1,689  \\
        CEBaB (\emph{all}) & & 11,728 & 1,673 & 1,689 \\
    \cmidrule{1-1} \cmidrule{3-5}
        CEBaB (\emph{paired}, \emph{human}) & & 19,684 & 3,898 & 3,958 \\
        CEBaB (\emph{paired}, \emph{sampling}) & & 74,574 & 3,898 & 3,958  \\
    \bottomrule
  \end{tabular}} 
 \caption{Dataset statistics.}
  \label{tab:cebab-stats}
\end{table*}

\subsection{Types of Approximate Counterfactual Pairs} \label{app:pairs-sampling}

Our approximate counterfactual training data comes in paired sentences of (\emph{original sentence}, \emph{approximate counterfactual sentence}). The approximate counterfactuals differs from their original counterparts in only one concept value. We consider approximate counterfactual pairs to be symmetric: we use both (\emph{original sentence}, \emph{approximate counterfactual sentence}) and (\emph{approximate counterfactual sentence}, \emph{original sentence}) as training pairs. 

\paragraph{Human-created Counterfactuals} CEBaB contains multiple counterfactual sentences for each original review. To achieve this, the dataset creators asked annotators to edit the original sentence to achieve a specified goal (e.g., `change the evaluation of the restaurant's food to negative'). These originals and corresponding edits form our \textit{human} pairs.

\paragraph{Metadata-sampled Counterfactuals} Human-created counterfactuals are not always available. With CEBaB, we simulate a second type of approximate counterfactuals by using metadata-guided heuristics: for a given \emph{original sentence}, we sample a counterfactual from the train set by matching concept labels while allowing only one label to be changed.

During training, we also consider \emph{null effect pairs} in our \emph{sampling} setup. These pairs resemble cases where our approximate counterfactual sentence is identical to the original sentence. When training our models on these pairs, we expect our models to predict the same counterfactual and factual output.

\subsection{Training Regimes} \label{app:training-setups}

\paragraph{\CPMIN{}} To train \CPMIN{}, we use the same model architecture as $\mathcal{N}$, and initialize it with the model weights using weights from $\mathcal{N}$. The maximum number of training epochs is set to 30 with a learning rate of $5e^{-5}$ and an effective batch size of 128. The learning rate linearly decays to $0$ over the 30 training epochs. We employ an early stopping strategy for $\text{COS}_{\texttt{ICaCE}}$ over the dev set for an interval of 50 steps with early stopping patience set to 20. We set the max sequence length to 128 and the dropout rate to $0.1$. We take a weighted sum of two objectives as the loss term for training \CPMHI{}. Specifically, we use $[w_{\text{Mimic}}, w_{\text{IN}}] = [1.0, 3.0]$. For the smoothed cross-entropy loss, we use a temperature of $2.0$.

\paragraph{\CPMHI{}} To train \CPMHI{}, we use the same model architecture as $\mathcal{N}$, and initialize it with the model weights using weights from $\mathcal{N}$. The maximum number of training epochs is set to 30 with a learning rate of $8e^{-5}$ and an effective batch size of 256. We use a higher learning rate of $0.001$ for the \texttt{LSTM} model as it enables quicker convergence. The learning rate linearly decays to $0$ over the 30 training epochs. We employ an early stopping strategy for $\text{COS}_{\texttt{ICaCE}}$ over the dev set for an interval of 10 steps with early stopping patience set to 20. We set the max sequence length to 128 and the dropout rate to $0.1$. We take a weighted sum of three objectives as the loss term for training \CPMHI{}. Specifically, we use $[w_{\text{Mimic}}, w_{\text{Multi}}, w_{\text{HI}}] = [1.0, 1.0, 3.0]$. In \appref{sec:iit-loss-terms}, we conduct a set of ablation studies to isolate the individual contributions from each objective. For the smoothed cross-entropy loss, we use a temperature of $2.0$.

Our models are all implemented in \texttt{PyTorch}~\citep{paszke2019pytorch} and using the \texttt{HuggingFace} library~\citep{wolf2019huggingface}. All of our results are aggregated over three distinct random seeds. To foster reproducibility, we will release our code repository and model artifacts to the public.

\begin{figure*}[tb]
    \centering
    \includegraphics[width=\textwidth]{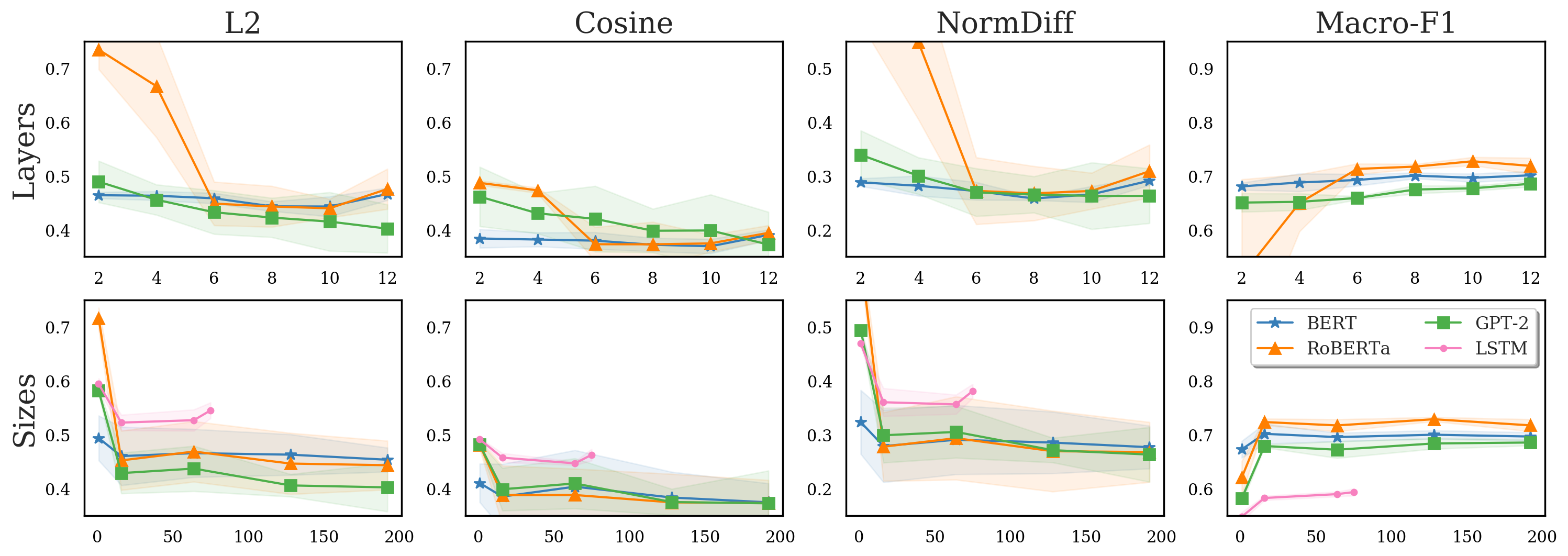}
    \caption{CEBaB scores for different intervention site locations and sizes for \CPMHI{}. The scores are measured in three different metrics on the test set for four different model architectures as a five-class sentiment classification task. Results averaged over three distinct seeds. Task performance as \texttt{Macro-F1} score is reported when applicable. Shaded areas outline $\pm$ SD. }
    \label{fig:interchange-site-location-and-size}
\end{figure*}

\subsection{Additional Baseline Results}\label{sec:ks-baselines}

\Tabref{tab:cebab-baselines} shows baselines adapted from \citet{abraham2022cebab}, which contains the present state-of-the-art explanation methods for the CEBaB benchmark. We report the best scores across these explanation methods in \tabref{tab:cebab-main}. These baselines are trained without using counterfactual data. Thus, we build additional baselines that use counterfactual data as shown in \tabref{tab:additional-baselines}. \texttt{S-Learner} is selected as the best performing models and included in \tabref{tab:cebab-main} for comparisons. The equations for the additional baselines are as follows:
\begin{align}
    \mathcal{E}_{\bbm}^{\text{approx}}(\ExplainerInput) &=
    \bbm(s^{\text{approx}}) - \bbm(\Factual)
    \\
    \mathcal{E}_{\bbm}^{\text{random}}(\ExplainerInput) &=
    \bbm(s^{\text{random}}) - \bbm(\Factual)
    \\
    \mathcal{E}_{\bbm}^{\text{CaCE}}(\ConceptEdit) &=    
        \frac{1}{\left|\mathcal{D}^{\ConceptEdit}\right|}
    \negthickspace
    \sum_{\ApprCounterfactualPairs \in \mathcal{D}^{\ConceptEdit}}
    \negthickspace 
    \! \! \! \!
    \left( 
        \bbm \left(\ApprCounterfactual \right)
        -
        \bbm \left(\Factual \right)
    \right)
    \\
    \mathcal{E}^{\text{ATE}}(\ConceptEdit) &=    
        \frac{1}{\left|\mathcal{D}^{\ConceptEdit}\right|}
    \negthickspace
    \sum_{\ApprCounterfactualPairs \in \mathcal{D}^{\ConceptEdit}}
    \negthickspace 
    \! \! \! \!
    \left( 
        f \left(\ApprCounterfactual \right)
        -
        f \left(\Factual \right)
    \right)
\end{align}

where $s^{\text{random}}$ is a randomly sampled training input, $s^{\text{approx}}$ is a training input sampled to match the concept-level labels of the true counterfactual under intervention $\ConceptEdit$, $\mathcal{D}^{\ConceptEdit}$ is the set of all approximate counterfactual training pairs that represent a $\ConceptEdit$ intervention, and $f$ is a look-up function that returns the ground-truth label associated with an input.

The signatures of $\mathcal{E}^{\text{ATE}}$ and $\mathcal{E}_{\bbm}^{\text{CaCE}}$ reflect that they are independent of the specific factual input $\Factual$ considered. Furthermore, $\mathcal{E}^{\text{ATE}}$ is independent of $\bbm{}$ given that this explainer only uses ground-truth training labels to estimate causal effects.

\begin{table*}[tp]
\centering
\resizebox{0.55\linewidth}{!}{%
  \centering
  \setlength{\tabcolsep}{4pt}
  \begin{tabular}{lrcccc}
\toprule
\multirow{1}{*}{Model} & \multirow{1}{*}{\MetricColName} & & \texttt{Approx}$^{\dagger}$ & \texttt{S-Learner}$^{\ddagger}$ & \texttt{INLP}$^{\mathsection}$ \\
\cmidrule{1-2} \cmidrule{4-6}
\multirow{3}{*}{\texttt{BERT}} 
& \MetricL & & 0.81 \ci{0.01} & 0.74 \ci{0.02} & 0.80 \ci{0.02} \\
& \MetricCos & & 0.61 \ci{0.01} & 0.63 \ci{0.01} & 0.59 \ci{0.03} \\
& \MetricNormDiff & & 0.44 \ci{0.01} & 0.54 \ci{0.02} & 0.73 \ci{0.02} \\
\cmidrule{1-2} \cmidrule{4-6}
\multirow{3}{*}{\texttt{RoBERTa}} 
& \MetricL & & 0.83 \ci{0.01} & 0.78 \ci{0.01} & 0.84 \ci{0.01} \\
& \MetricCos & & 0.60 \ci{0.01} & 0.64 \ci{0.01} & 0.58 \ci{0.01} \\
& \MetricNormDiff & & {0.45} \ci{0.00} & 0.59 \ci{0.01} & 0.81 \ci{0.01} \\
\cmidrule{1-2} \cmidrule{4-6}
\multirow{3}{*}{\texttt{GPT-2}} 
& \MetricL & & 0.72 \ci{0.02} & 0.60 \ci{0.02} & 0.72 \ci{0.01} \\
& \MetricCos & & 0.59 \ci{0.01} & 0.59 \ci{0.01} & 1.00 \ci{0.00} \\
& \MetricNormDiff & & 0.41 \ci{0.01} & 0.40 \ci{0.01} & 0.58 \ci{0.03} \\
\cmidrule{1-2} \cmidrule{4-6}
\multirow{3}{*}{\texttt{LSTM}} 
& \MetricL & & 0.86 \ci{0.01} & 0.73 \ci{0.01} & 0.79 \ci{0.01} \\
& \MetricCos & & 0.64 \ci{0.01} & 0.64 \ci{0.01} & 0.74 \ci{0.02} \\
& \MetricNormDiff & & 0.50 \ci{0.01} & 0.53 \ci{0.01} & 0.60 \ci{0.01} \\
\bottomrule
\end{tabular}}
  \caption{CEBaB scores measured in three different metrics on the test set for four different model architectures as a five-class sentiment classification task. Results are adapted from \citet{abraham2022cebab}. \textbf{Lower is better}; standard deviations over 5 distinct seeds in parentheses. Results are aggregated over all aspects and all directional concept label changes. Details about these evaluation metrics can be found in~\Secref{sec:cpm-eval}. Results are based on $^{\dagger}$\citet{abraham2022cebab}, $^{\ddagger}$ \citet{kunzel2019metalearners}, and $^{\mathsection}$\citet{ravfogel_null_2020}.}
  \label{tab:cebab-baselines}
\end{table*}

\begin{table*}[tp]
\centering
\resizebox{1.0\linewidth}{!}{%
  \centering
  \setlength{\tabcolsep}{4pt}
  \begin{tabular}{lrcccccccccc}
\toprule
& & & \multicolumn{4}{c}{\emph{sampled counterfactuals}} & & \multicolumn{4}{c}{\emph{human-created counterfactuals}} \\
& & & & \texttt{ATE-} & \texttt{CaCE-} & & & & \texttt{ATE-} & \texttt{CaCE-} & \\
\multirow{1}{*}{Model} & \multirow{1}{*}{\MetricColName} & & \texttt{Approx} & \texttt{Explainer} & \texttt{Explainer} & \texttt{Random} & & \texttt{Approx} & \texttt{Explainer} & \texttt{Explainer} & \texttt{Random} \\
\cmidrule{1-2} \cmidrule{4-7} \cmidrule{9-12}
\multirow{3}{*}{\texttt{BERT}} 
& \MetricL &  & 0.81 \ci{0.01} & 0.81 \ci{0.02} & 0.81 \ci{0.02} & 0.84 \ci{0.02} &  & 0.79 \ci{0.02} & 0.81 \ci{0.02} & 0.80 \ci{0.02} & 0.84 \ci{0.01} \\
& \MetricCos &  & 0.60 \ci{0.00} & 0.72 \ci{0.01} & 0.72 \ci{0.01} & 0.53 \ci{0.00} &  & 0.56 \ci{0.01} & 0.69 \ci{0.01} & 0.69 \ci{0.01} & 0.53 \ci{0.00} \\
& \MetricNormDiff &  & 0.44 \ci{0.01} & 0.62 \ci{0.02} & 0.62 \ci{0.02} & 0.55 \ci{0.02} &  & 0.43 \ci{0.01} & 0.62 \ci{0.02} & 0.64 \ci{0.02} & 0.54 \ci{0.02} \\
\cmidrule{1-2} \cmidrule{4-7} \cmidrule{9-12}
\multirow{3}{*}{\texttt{RoBERTa}} 
& \MetricL &  & 0.83 \ci{0.01} & 0.85 \ci{0.00} & 0.85 \ci{0.00} & 0.87 \ci{0.00} &  & 0.81 \ci{0.01} & 0.85 \ci{0.00} & 0.84 \ci{0.00} & 0.87 \ci{0.00} \\
& \MetricCos &  & 0.61 \ci{0.01} & 0.73 \ci{0.00} & 0.73 \ci{0.01} & 0.53 \ci{0.00} &  & 0.57 \ci{0.01} & 0.70 \ci{0.00} & 0.70 \ci{0.00} & 0.53 \ci{0.00} \\
& \MetricNormDiff &  & 0.46 \ci{0.01} & 0.67 \ci{0.00} & 0.67 \ci{0.00} & 0.58 \ci{0.00} &  & 0.44 \ci{0.01} & 0.67 \ci{0.00} & 0.68 \ci{0.00} & 0.59 \ci{0.00} \\
\cmidrule{1-2} \cmidrule{4-7} \cmidrule{9-12}
\multirow{3}{*}{\texttt{GPT-2}} 
& \MetricL &  & 0.72 \ci{0.02} & 0.69 \ci{0.01} & 0.68 \ci{0.01} & 0.76 \ci{0.00} &  & 0.72 \ci{0.01} & 0.68 \ci{0.01} & 0.68 \ci{0.01} & 0.76 \ci{0.00} \\
& \MetricCos &  & 0.59 \ci{0.00} & 0.67 \ci{0.00} & 0.67 \ci{0.00} & 0.56 \ci{0.00} &  & 0.57 \ci{0.00} & 0.66 \ci{0.00} & 0.65 \ci{0.00} & 0.56 \ci{0.00} \\
& \MetricNormDiff &  & 0.40 \ci{0.01} & 0.48 \ci{0.01} & 0.49 \ci{0.01} & 0.47 \ci{0.00} &  & 0.40 \ci{0.00} & 0.49 \ci{0.01} & 0.50 \ci{0.01} & 0.47 \ci{0.01} \\
\cmidrule{1-2} \cmidrule{4-7} \cmidrule{9-12}
\multirow{3}{*}{\texttt{LSTM}} 
& \MetricL &  & 0.87 \ci{0.00} & 0.78 \ci{0.00} & 0.78 \ci{0.00} & 0.85 \ci{0.00} &  & 0.85 \ci{0.01} & 0.78 \ci{0.00} & 0.76 \ci{0.00} & 0.84 \ci{0.00} \\
& \MetricCos &  & 0.65 \ci{0.00} & 0.71 \ci{0.00} & 0.71 \ci{0.00} & 0.57 \ci{0.00} &  & 0.61 \ci{0.00} & 0.69 \ci{0.00} & 0.68 \ci{0.00} & 0.56 \ci{0.00} \\
& \MetricNormDiff &  & 0.50 \ci{0.00} & 0.59 \ci{0.00} & 0.59 \ci{0.00} & 0.55 \ci{0.00} &  & 0.49 \ci{0.00} & 0.59 \ci{0.00} & 0.61 \ci{0.00} & 0.55 \ci{0.00} \\
\bottomrule
\end{tabular}}
  \caption{CEBaB scores for additional baselines we considered. CEBaB scores are measured in three different metrics on the test set for four different model architectures as a five-class sentiment classification task. \textbf{Lower is better}. Results averaged over three distinct seeds, standard deviations in parentheses. Details about these evaluation metrics can be found in~\Secref{sec:cpm-eval}.}
  \label{tab:additional-baselines}
\end{table*}

\subsection{Intervention Site Location and Size}\label{sec:iit-location}

Previous work shows that neurons in different layers and groups can encode different high-level concepts~\citep{vig2020causal, koh2020concept}. \CPMHI{} pushes concept-related information to localize at the targeted intervention site (the aligned neural representations for each concept). In this section, we investigate how the location and the size of the intervention site impact \CPMHI{} performance. We use the optimal location and size found in this study for other results presented in this paper.

\paragraph{Location} For Transformer-based models, we vary the location of the intervention site by intervening on the ``[CLS]'' token embedding layer $l$. Specifically, we set $l = \{2, 4, 6, 8, 10, 12\}$. We skip this experiment for non-Transformer-based model (i.e., \texttt{LSTM}) since it only contains a single sentence embedding.

As shown in the top panel of \figref{fig:interchange-site-location-and-size}, intervention location significantly affects \CPMHI{} performance. Our results show that layer 10 for \texttt{BERT}, layer 8 for \texttt{RoBERTa}, and layer 12 for \texttt{GPT-2} lead to the best performance. This suggests layers have different efficacy in terms of information localization. Our results also show that intervening with deeper layers tends to provide better performance. However, for both \texttt{BERT} and \texttt{RoBERTa}, intervening on the last layer results in a slightly worse performance compared to earlier layers. This suggests that leaving Transformer blocks after the intervention site helps localized information to be processed by the neural network.

\paragraph{Size} For Transformer-based models, we change the size of the intervention site $d_{c}$ for each concept. Specifically, we set $d_{c} = \{1, 16, 64, 128, 192\}$. For instance when $d_{c} = 1$, we use a single dimension of the ``[CLS]'' token embedding to represent each concept, starting from the first dimension of the vector. For our non-Transformer-based model (\texttt{LSTM}), we intervene on the attention-gated sentence embedding whose dimension size is set to 300. Accordingly, we set $d_{c} = \{1, 16, 64, 75\}$. 

As shown in \figref{fig:interchange-site-location-and-size}, larger intervention sites lead to better performance for all Transformer-based models. For \texttt{LSTM}, we find that the optimal size is the second largest one instead. On the other hand, our results suggest that the performance gain from the increase of size diminishes as we increase the size for all model architectures. 

\begin{table*}[tp]
\centering
\resizebox{0.75\linewidth}{!}{%
  \centering
  \setlength{\tabcolsep}{4pt}
  \begin{tabular}{llccccc}
\toprule
Model & Ablation & & \MetricL & \MetricCos & \MetricNormDiff & \texttt{Macro-F1} \\
\cmidrule{1-2} \cmidrule{4-7}
\multirow{5}{1.5cm}{\texttt{BERT}}
& \textbf{\CPMHI} & & 0.45 \ci{0.02} & 0.36 \ci{0.03} & 0.27 \ci{0.04} & 0.69 \ci{0.01} \\
& ${}-\mathcal{L}_{\text{Multi}}$ & & 0.47 \ci{0.04} & 0.38 \ci{0.04} & 0.30 \ci{0.07} & 0.69 \ci{0.01} \\
& ${}-\mathcal{L}_{\text{HI}}$ & & 0.79 \ci{0.02} & 0.60 \ci{0.03} & 0.64 \ci{0.02} & 0.60 \ci{0.08}\\
& ${}+\emph{random init}$ & & 0.81 \ci{0.02} & 0.52 \ci{0.00} & 0.55 \ci{0.02} & 0.08 \ci{0.02} \\
& ${}+\emph{no training}$ & & 0.80 \ci{0.02} & 0.86 \ci{0.04} & 0.76 \ci{0.02} & 0.70 \ci{0.01} \\
\cmidrule{1-2} \cmidrule{4-7}
\multirow{5}{1.5cm}{\texttt{RoBERTa}}
& \textbf{\CPMHI} & & 0.47 \ci{0.03} & 0.39 \ci{0.03} & 0.29 \ci{0.05} & 0.71 \ci{0.00} \\
& ${}-\mathcal{L}_{\text{Multi}}$ & & 0.49 \ci{0.05} & 0.41 \ci{0.05} & 0.32 \ci{0.06} & 0.70 \ci{0.00} \\
& ${}-\mathcal{L}_{\text{HI}}$ & & 0.81 \ci{0.00} & 0.53 \ci{0.02} & 0.63 \ci{0.01} & 0.39 \ci{0.06} \\
& ${}+\emph{random init}$ & & 0.85 \ci{0.00} & 0.51 \ci{0.00} & 0.59 \ci{0.01} & 0.06 \ci{0.00} \\
& ${}+\emph{no training}$ & & 0.84 \ci{0.01} & 0.93 \ci{0.05} & 0.83 \ci{0.00} & 0.70 \ci{0.00} \\
\cmidrule{1-2} \cmidrule{4-7}
\multirow{5}{1.5cm}{\texttt{GPT-2}}
& \textbf{\CPMHI} & & 0.41 \ci{0.04} & 0.39 \ci{0.05} & 0.27 \ci{0.05} & 0.68 \ci{0.00} \\
& ${}-\mathcal{L}_{\text{Multi}}$ & & 0.43 \ci{0.03} & 0.41 \ci{0.05} & 0.29 \ci{0.04} & 0.67 \ci{0.00} \\
& ${}-\mathcal{L}_{\text{HI}}$ & & 0.66 \ci{0.01} & 0.58 \ci{0.04} & 0.49 \ci{0.01} & 0.58 \ci{0.04} \\
& ${}+\emph{random init}$ & & 0.73 \ci{0.00} & 0.54 \ci{0.00} & 0.47 \ci{0.01} & 0.16 \ci{0.00} \\
& ${}+\emph{no training}$ & & 0.65 \ci{0.00} & 0.61 \ci{0.00} & 0.57 \ci{0.02} & 0.65 \ci{0.00} \\
\cmidrule{1-2} \cmidrule{4-7}
\multirow{5}{1.5cm}{\texttt{LSTM}}
& \textbf{\CPMHI} & & 0.54 \ci{0.01} & 0.46 \ci{0.01} & 0.36 \ci{0.00} & 0.59 \ci{0.01} \\
& ${}-\mathcal{L}_{\text{Multi}}$ & & 0.56 \ci{0.02} & 0.47 \ci{0.02} & 0.41 \ci{0.02} & 0.59 \ci{0.01} \\
& ${}-\mathcal{L}_{\text{HI}}$ & & 0.73 \ci{0.00} & 0.64 \ci{0.02} & 0.59 \ci{0.00} & 0.59 \ci{0.01} \\
& ${}+\emph{random init}$ & & 0.82 \ci{0.00} & 0.55 \ci{0.00} & 0.55 \ci{0.00} & 0.13 \ci{0.04} \\
& ${}+\emph{no training}$ & & 0.73 \ci{0.01} & 0.74 \ci{0.00} & 0.59 \ci{0.01} & 0.60 \ci{0.01} \\
\bottomrule
\end{tabular}}
  \caption{Ablation study of our $\CPMHI{}$ method trained with \emph{human} approximate counterfactual strategy. CEBaB scores measured in three different metrics on the test set for four different model architectures as a five-class sentiment classification task. \textbf{Lower is better}. Results averaged over three distinct seeds, standard deviations in parentheses.}
  \label{tab:objective-ablation}
\end{table*}

\begin{table*}[tp]
\centering
\resizebox{0.85\linewidth}{!}{%
  \centering
  \setlength{\tabcolsep}{4pt}
  \begin{tabular}{lrcccccccc}
\toprule
& & & \multicolumn{3}{c}{\emph{sampled counterfactuals}} & & \multicolumn{3}{c}{\emph{human-created counterfactuals}} \\
& & & & \texttt{Random} & \texttt{Probe-based} & & & \texttt{Random} & \texttt{Probe-based} \\
\multirow{1}{*}{Model} & \multirow{1}{*}{\MetricColName} & & \textbf{\CPMHI} & \texttt{Source} & \texttt{Source} & & \textbf{\CPMHI} & \texttt{Source} & \texttt{Source} \\
\cmidrule{1-2} \cmidrule{4-6} \cmidrule{8-10}
\multirow{3}{*}{\texttt{BERT}} 
& \MetricL &  & {0.60} \ci{0.01} & 0.74 \ci{0.03} & 0.61 \ci{0.01} &  & 0.45 \ci{0.03} & 0.70 \ci{0.03} & 0.43 \ci{0.02} \\
& \MetricCos &  & 0.45 \ci{0.00} & 0.53 \ci{0.01} & 0.45 \ci{0.00} &  & 0.36 \ci{0.04} & 0.59 \ci{0.04} & 0.35 \ci{0.01} \\
& \MetricNormDiff &  & 0.38 \ci{0.00} & 0.54 \ci{0.02} & 0.39 \ci{0.01} &  & 0.27 \ci{0.01} & 0.53 \ci{0.01} & 0.25 \ci{0.02} \\
\cmidrule{1-2} \cmidrule{4-6} \cmidrule{8-10}
\multirow{3}{*}{\texttt{RoBERTa}} 
& \MetricL &  & 0.67 \ci{0.02} & 0.79 \ci{0.01} & 0.66 \ci{0.02} &  & 0.47 \ci{0.03} & 0.72 \ci{0.01} & 0.44 \ci{0.01} \\
& \MetricCos &  & 0.47 \ci{0.00} & 0.52 \ci{0.01} & 0.46 \ci{0.01} &  & 0.39 \ci{0.03} & 0.57 \ci{0.03} & 0.37 \ci{0.01} \\
& \MetricNormDiff &  & 0.45 \ci{0.03} & 0.59 \ci{0.00} & 0.44 \ci{0.03} &  & 0.29 \ci{0.05} & 0.55 \ci{0.01} & 0.25 \ci{0.01} \\
\cmidrule{1-2} \cmidrule{4-6} \cmidrule{8-10}
\multirow{3}{*}{\texttt{GPT-2}} 
& \MetricL &  & 0.51 \ci{0.01} & 0.65 \ci{0.02} & 0.51 \ci{0.02} &  & 0.41 \ci{0.04} & 0.58 \ci{0.03} & 0.39 \ci{0.02} \\
& \MetricCos &  & 0.46 \ci{0.00} & 0.55 \ci{0.01} & 0.46 \ci{0.01} &  & 0.39 \ci{0.05} & 0.56 \ci{0.02} & 0.37 \ci{0.01} \\
& \MetricNormDiff &  & 0.30 \ci{0.00} & 0.46 \ci{0.01} & 0.31 \ci{0.01} &  & 0.27 \ci{0.05} & 0.44 \ci{0.01} & 0.25 \ci{0.01} \\
\cmidrule{1-2} \cmidrule{4-6} \cmidrule{8-10}
\multirow{3}{*}{\texttt{LSTM}} 
& \MetricL &  & 0.64 \ci{0.02} & 0.76 \ci{0.01} & 0.65 \ci{0.02} &  & 0.54 \ci{0.01} & 0.69 \ci{0.03} & 0.55 \ci{0.00} \\
& \MetricCos &  & 0.50 \ci{0.01} & 0.57 \ci{0.01} & 0.50 \ci{0.01} &  & 0.46 \ci{0.00} & 0.58 \ci{0.01} & 0.46 \ci{0.01} \\
& \MetricNormDiff &  & 0.41 \ci{0.01} & 0.54 \ci{0.01} & 0.41 \ci{0.02} &  & 0.36 \ci{0.00} & 0.52 \ci{0.00} & 0.38 \ci{0.01} \\
\bottomrule
\end{tabular}}
  \caption{Ablation study of our $\CPMHI{}$ method for different \emph{source} input $\VarSourceInput$ sampling strategies at inference time. CEBaB scores measured in three different metrics on the test set for four different model architectures as a five-class sentiment classification task. \textbf{Lower is better}. Results averaged over three distinct seeds, standard deviations in parentheses.}
  \label{tab:additional-baselines-source-labels}
\end{table*}

\subsection{Ablation Study of \CPMHI{}}\label{sec:iit-loss-terms}

\citet{geiger-etal-2021-iit} show that training with a multi-task objective helps IIT to improve generalizability. In this experiment, we aim to investigate whether the multi-task objective we added for \CPMHI{} plays an important role in achieving good performance. Specifically, we conduct two ablation studies: removing the multi-task objective by setting $w_{\text{Multi}} = 0.0$, and removing the IIT objective by setting  $w_{\text{HI}} = 0.0$.

\Tabref{tab:objective-ablation} shows our results, which demonstrate that the IIT objective is the main factor that drives \CPMHI{} performance. Our results also suggest that the multi-task objective brings relatively small but consistent performance gains. Overall, our findings corroborate those of  \citet{geiger-etal-2021-iit} and provide concrete evidence that the combination of two objectives always results in the best-performing explanation methods across all model architectures.

Additionally, we explore two baselines for \CPMHI{}. Firstly, we randomly initialize the weights of \CPMHI{}. Secondly, we take the original black-box model as our \CPMHI{}. Compared to the results in \tabref{tab:cebab-main}, these two baselines fail catastrophically, suggesting the importance of our IIT paradigm.

As mentioned in \secref{sec:CPM}, we sample a source input $\SourceInput$ from the train set as any input $\VarInput$ that has $\ConceptAssign$ to estimate the counterfactual output. Furthermore, we explore two additional sampling strategies. First, we create a baseline where we randomly sample a source input from the train without any concept label matching. Second, we sample a source input from the train  set using the predicted concept label of our multi-task probe, instead of the true concept label from the dataset.

As shown in \tabref{tab:additional-baselines-source-labels}, the quality of our source inputs impact our performance significantly. For instance, when sampling source input at random, \CPMHI{} fails catastrophically for all evaluation metrics. On the other hand, when we sampling source based on the predicted labels using the multi-task probe, \CPMHI{} maintains its performance.

\subsection{GPT-3 Generation Process} \label{app:gpt-3}

\begin{figure*}[t!]
    \centering
    
    \fbox{
    \begin{minipage}{0.95\textwidth}
    \texttt{\textcolor{gray}{Make the following restaurant reviews include POSITIVE mentions of SERVICE.}}\\
    
    \texttt{\textcolor{gray}{Original: I had two casual dinners at State \& Lake and three lunches. The food was great but the service  was lacking. Everything was delicious. The interior is questionable, but not intrusive.}} \\
    
    \texttt{\textcolor{gray}{POSITIVE mentions of SERVICE: I had two casual dinners at State \& Lake and three lunches. The food and the service were always great. Everything was delicious. The interior is questionable, but not intrusive.}} \\
    
    \texttt{\textcolor{gray}{Original: Food was excellent, but the service was not very attentive. Noise level was extremely high due to close proximity of tables and poor acoustics.}} \\
    
    \texttt{\textcolor{gray}{POSITIVE mentions of SERVICE: Food and service was excellent. Noise level was extremely high due to close proximity of tables and poor acoustics.}} \\
    
    \texttt{\textcolor{gray}{Original: Great food, poor and very snobbish service.}} \\
    
    \texttt{\textcolor{gray}{POSITIVE mentions of SERVICE: Great food, very good service.}} \\
    
    \texttt{\textcolor{gray}{Original: My dining experince was excellent! However, the server was not nice.}} \\
    
    \texttt{\textcolor{gray}{POSITIVE mentions of SERVICE: My dining experince was excellent!}} \\
    
    \texttt{\textcolor{gray}{Original: Hae been here a few times and it is just okay - Entrees and wine list a bit pricey for what it is, inattentive staff.}} \\
    
    \texttt{\textcolor{gray}{POSITIVE mentions of SERVICE: Hae been here a few times and it is just okay - Entrees and wine list a bit pricey for what it is. Food comes out on time.}} \\
    
    \texttt{\textcolor{gray}{Original: Tables fairly close together, mushroom appetiser very good, pork entree fair, chicken good. The service was terrible.}} \\
    
    \texttt{\textcolor{gray}{POSITIVE mentions of SERVICE: Tables fairly close together, mushroom appetiser very good, pork entree fair, chicken good. The service was great however.}} \\
    
    \texttt{\textcolor{gray}{Original: Service was very poor with the server unresponsive and misinformed on all requests. The food was very good with a good selection of entrees. The ambiance was romantic with a quiet excellence.}} \\
    
    \texttt{\textcolor{gray}{POSITIVE mentions of SERVICE:}} \texttt{\textbf{Service was very good with the server attentive and responsive on all requests. The food was very good with a good selection of entrees. The ambiance was romantic with a quiet excellence.}}
    \end{minipage}}
    
    \caption{Example \texttt{GPT-3} prompt (gray) and \texttt{GPT-3} completion (bold). Note that all original examples convey the same sentiment towards service ($\VarConcept = \text{negative}$) and same overall sentiment ($y=\text{neutral}$), and that the counterfactual examples are all edited such that the sentiment towards service is the same ($\VarConceptEdit = \text{positive}$).}
    
    \label{fig:gpt3-example}
\end{figure*}

\begin{figure*}[t!]
    \centering
    
    \fbox{
    \begin{minipage}{0.95\textwidth}
    \texttt{\textcolor{gray}{Make the following restaurant reviews include POSITIVE mentions of SERVICE.}}\\
    
    \texttt{\textcolor{gray}{Original: Been here several times. Always a winner, except for the tasteless food!}} \\
    
    \texttt{\textcolor{gray}{POSITIVE mentions of SERVICE: I was very disappointed in the food but we did not wait long for each course and or waiter was very pleasant.}} \\
    
    \texttt{\textcolor{gray}{Original: food was decent but not great.}} \\
    
    \texttt{\textcolor{gray}{POSITIVE mentions of SERVICE: Lovely evening - good service and wonderful food. Perfect for fresh fish fans}} \\
    
    \texttt{\textcolor{gray}{Original: The restaurant was empty when we arrived, reservation not necessary? Wine list limited. Food was bland, presentation was very well done. I would not eat here again.}} \\
    
    \texttt{\textcolor{gray}{POSITIVE mentions of SERVICE: Abby provided the best service that we've had after probably two dozen visits. No thank you for making the risotto cake at lunch....Two Stars!}} \\
    
    \texttt{\textcolor{gray}{Original: A terrible place for lunch or dinner. All the food is excellent with top notch ingredients}} \\
    
    \texttt{\textcolor{gray}{POSITIVE mentions of SERVICE: Excellent Valentine's menu. Excellent service and food. Would recommend this restaurant and will return.}} \\
    
    \texttt{\textcolor{gray}{Original: The food was average for the cost. My husband and I were so excited to visit Bobby Flay's restraunt and were really disappointed. The food was average at best.}} \\
    
    \texttt{\textcolor{gray}{POSITIVE mentions of SERVICE:}} \texttt{\textbf{The service was amazing and the food was alright.}}
    \end{minipage}}
    
    \caption{Example \texttt{GPT-3} prompt (gray) and \texttt{GPT-3} completion (bold). Note that all original examples convey the same sentiment towards service ($\VarConcept = \text{unknown}$) and same overall sentiment ($y=\text{negative}$), and that the counterfactual examples are all metadata-sampled such that the sentiment towards service is the same ($\VarConceptEdit = \text{positive}$).}
    
    \label{fig:gpt3-approximate-example}
\end{figure*}

We use the 175B parameter \texttt{davinci} \texttt{GPT-3} model \citep{brown_language_2020} as a few-shot learner to generate approximate counterfactual data. Let $\Factual$ be a review text with an original value $\VarConcept$ for the mediating concept $\VarConceptAbstractSingle$ and an overall review sentiment $y$ (e.g., a restaurant review which is \textit{negative} about the \textit{service}, and felt \textit{neutral} about their overall dining experience), and let $\VarConceptEdit$ be the target value of $\VarConceptAbstractSingle$, for which we would like to create a counterfactual review (e.g., change the text to become \textit{positive} about the mediating concept \textit{service}). In order to use \texttt{GPT-3} as an $n$-shot learner, we sample $n = 6$ approximate counterfactual pairs $\GPTApprCounterfactualPairs$, where $\VarSourceInputAbstract_{\UVEditWorld}$ shares with $\Factual$ the same value $\VarConcept$ for $\VarConceptAbstractSingle$ and the same overall sentiment, and the counterfactual review $\ac{\VarSourceInputAbstract}_{\UVEditWorld}^{\ConceptEdit}$ has the target value $\VarConceptEdit$ for $\VarConceptAbstractSingle$. We prompt the model with these pairs, and we also include the original review $\Factual$. We then collect the text completed by \texttt{GPT-3} as the \texttt{GPT-3} counterfactual review. An example for this $n$-shot prompt and completion is in \figref{fig:gpt3-example}. In addition, we also prompt \texttt{GPT-3} with pairs of original reviews and metadata-sampled counterfactuals, and generate another set of \texttt{GPT-3} counterfactual review for comparison. We sample $n = 4$ approximate counterfactual pairs in this case. An example of metadata-sampled counterfactual generation with GPT-3 can be seen in \figref{fig:gpt3-approximate-example}.

For each few-shot learning prompt, we insert an initial string of the form of ``Make the following restaurant reviews include $c'$ mentions of $C_i$.'', where $c'$ is expressed as one of \{``POSITIVE'', ``NEGATIVE'', ``NOT'' \} (``NOT'' corresponds to making the review be unknown regarding the concept $C_i$) and $C_i$ is one of \{``AMBIANCE'', ``FOOD'', ``NOISE'', ``SERVICE''\}. We sample using a temperature of 0.9, without any frequency or presence penalties (since we expect the counterfactual review to be  similar to the original review). In preliminary experimentation, we found that capitalizing the mediating concept and target value results and inserting line breaks between examples made for better completions, although there is room for future research in this area. 

We used the OpenAI API to access \texttt{GPT-3}. At the current price rate of \$0.02 per 1,000 tokens, the total cost of creating our counterfactuals (around 4,000 examples) was approximately \$50 per approximate counterfactuals creation strategy.

\subsection{Integrated Gradients}\label{app:ig-method}

We adapt the Integrated Gradients (IG) method of \citet{sundararajan17a} to qualitatively assess whether \CPMHI{} learned explainable representations of mediated concepts at its intervention sites. The IG algorithm computes the average gradient from the model output to its input by incrementally interpolating from a ``blank'' input $x'$ (consisting only of ``[PAD]'' tokens) to the original input $x$. \Eqnref{eqn:integrated-gradients} is the integrated gradients equation originally proposed in \citet{sundararajan17a}, applied to a CPM model $\mathcal{P}$ on input $x$.
\begin{equation}
    \text{IntegratedGrads}_j(x) = 
    (x_j - x_j')
    \cdot 
    \int_{\alpha=0}^1 
    \frac{
        \partial \mathcal{P}(x' + \alpha \cdot (x - x'))
    }{
        \partial x_j
    } \partial \alpha
    \label{eqn:integrated-gradients}
\end{equation}
Here, $\frac{\partial \mathcal{P}(x)}{\partial x_j}$ is the derivative of $\mathcal{P}$ on the $j$th dimension of $x$.

In our implementation of IG, we wish to show the per-token attribution of input $x$ on the model's final output $\mathcal{P}(x)$, \textit{mediated by} the hidden representation of a concept in $\mathcal{P}$. That is, we'd like to ask, ``What is the effect of the word `delicious' in the input on the model's output, when we restrict our focus only on the model's representation of the concept \textit{food}?'' 

To answer this question, we compute the gradient of the model output $\mathcal{P}(x)$ with respect to the input $x$ but restrict the gradient to flow through the intervention site for a particular concept. This allows us to capture the per-token attribution of the model's final output (whether particular words contributed to a \textit{positive}, \textit{negative}, or \textit{neutral} sentiment prediction), mediated by the concept that is represented by the specified intervention site. For example, in \tabref{tab:ig}, we can see that ``delicious'' has a positive attribution to the output of the model when we focus on its representation of the concept \textit{food}. 

Formally, consider a trained CPM model $\mathcal{P}$, an input $\VarInput$ and mediating concept $\VarConceptAbstractSingle$. Let $\HiddenForConcept$ be the activation of $\mathcal{P}$ at the intervention site for $\VarConceptAbstractSingle$. We define the gradient of $\mathcal{P}(\VarInput)$ along dimension $j$, \textit{mediated by} $\VarConceptAbstractSingle$, as 

\begin{equation}
    \frac{
        \partial \mathcal{P}(\VarInput)
    }{
        \partial \VarInput_j
    } 
    \text{ mediated by } \VarConceptAbstractSingle = 
    \frac{
        \partial \mathcal{P}(\VarInput)
    }{
        \partial \HiddenForConcept
    }
    \cdot 
    \frac{
        \partial \HiddenForConcept
    }{
        \partial \VarInput_j
    }.
    \label{eqn:mediated-gradient}
\end{equation}

\Eqnref{eqn:mediated-gradient} restricts the gradient to only flow through the hidden representation of the concept along which we'd like to interpret our model.

We integrate these mediated gradients over a straight path between input $x$ and baseline $x'$, analogous to \Eqnref{eqn:integrated-gradients}. We implement our IG method using \texttt{CaptumAI} library.\footnote{\url{https://captum.ai/}} We use the default parameters for our runs with number of iterations set to 50, and we set the integral method as \texttt{gausslegendre}. We set the \texttt{multiply-by-inputs} flag to \texttt{True}. To visualize individual word importance, we conduct $z$-score normalization of attribution scores over input tokens per each concept, and then linearly scale scores between [$-1$, $+1$].

\begin{table*}[tp]
\centering
\resizebox{0.99\linewidth}{!}{%
  \centering
  \setlength{\tabcolsep}{4pt}
  \begin{tabular}[c]{lllcc}
\toprule
Model & Predicted & Concept & Score & Word Importance \\
\midrule
\multirow{5}{*}{\texttt{Finetuned}} & \multirow{5}{*}{\emph{neutral}}
& ambiance & $+0.03$ & \colorbox{red!10}{\strut [CLS]}\colorbox{red!36}{\strut the}\colorbox{green!37}{\strut music}\colorbox{green!12}{\strut was}\colorbox{red!3}{\strut too}\colorbox{green!19}{\strut loud}\colorbox{red!20}{\strut ,}\colorbox{red!23}{\strut and}\colorbox{green!0}{\strut the}\colorbox{red!25}{\strut decorations}\colorbox{red!16}{\strut were}\colorbox{red!11}{\strut taste}\colorbox{green!41}{\strut \#\#less}\colorbox{red!32}{\strut ,}\colorbox{green!34}{\strut but}\colorbox{green!16}{\strut they}\colorbox{red!37}{\strut had}\colorbox{green!67}{\strut friendly}\colorbox{red!31}{\strut waiter}\colorbox{red!50}{\strut \#\#s}\colorbox{red!11}{\strut and}\colorbox{green!100}{\strut delicious}\colorbox{red!7}{\strut pasta}\colorbox{red!10}{\strut [SEP]} \\
& & food & $+0.11$ & \colorbox{red!14}{\strut [CLS]}\colorbox{green!14}{\strut the}\colorbox{red!38}{\strut music}\colorbox{red!43}{\strut was}\colorbox{green!62}{\strut too}\colorbox{green!33}{\strut loud}\colorbox{red!8}{\strut ,}\colorbox{red!18}{\strut and}\colorbox{green!35}{\strut the}\colorbox{red!5}{\strut decorations}\colorbox{red!11}{\strut were}\colorbox{red!11}{\strut taste}\colorbox{red!21}{\strut \#\#less}\colorbox{red!11}{\strut ,}\colorbox{red!3}{\strut but}\colorbox{green!12}{\strut they}\colorbox{red!37}{\strut had}\colorbox{green!100}{\strut friendly}\colorbox{red!51}{\strut waiter}\colorbox{green!10}{\strut \#\#s}\colorbox{green!2}{\strut and}\colorbox{green!68}{\strut delicious}\colorbox{red!39}{\strut pasta}\colorbox{red!24}{\strut [SEP]} \\
& & noise & $+0.04$ & \colorbox{green!6}{\strut [CLS]}\colorbox{green!24}{\strut the}\colorbox{red!19}{\strut music}\colorbox{red!3}{\strut was}\colorbox{green!11}{\strut too}\colorbox{green!13}{\strut loud}\colorbox{red!11}{\strut ,}\colorbox{red!4}{\strut and}\colorbox{green!5}{\strut the}\colorbox{red!6}{\strut decorations}\colorbox{red!10}{\strut were}\colorbox{red!12}{\strut taste}\colorbox{red!0}{\strut \#\#less}\colorbox{red!14}{\strut ,}\colorbox{green!4}{\strut but}\colorbox{green!12}{\strut they}\colorbox{red!51}{\strut had}\colorbox{green!100}{\strut friendly}\colorbox{red!23}{\strut waiter}\colorbox{red!34}{\strut \#\#s}\colorbox{red!58}{\strut and}\colorbox{green!72}{\strut delicious}\colorbox{green!5}{\strut pasta}\colorbox{red!3}{\strut [SEP]} \\
& & service & $+0.26$ & \colorbox{red!8}{\strut [CLS]}\colorbox{green!87}{\strut the}\colorbox{red!33}{\strut music}\colorbox{green!38}{\strut was}\colorbox{green!11}{\strut too}\colorbox{red!9}{\strut loud}\colorbox{red!54}{\strut ,}\colorbox{red!2}{\strut and}\colorbox{red!14}{\strut the}\colorbox{red!33}{\strut decorations}\colorbox{red!58}{\strut were}\colorbox{red!39}{\strut taste}\colorbox{green!49}{\strut \#\#less}\colorbox{red!14}{\strut ,}\colorbox{green!61}{\strut but}\colorbox{green!2}{\strut they}\colorbox{green!0}{\strut had}\colorbox{red!7}{\strut friendly}\colorbox{red!42}{\strut waiter}\colorbox{red!56}{\strut \#\#s}\colorbox{green!51}{\strut and}\colorbox{green!100}{\strut delicious}\colorbox{red!9}{\strut pasta}\colorbox{red!17}{\strut [SEP]} \\
\midrule
\multirow{5}{*}{\textbf{\CPMHI}} & \multirow{5}{*}{\emph{neutral}}
& ambiance & $-0.61$ & \colorbox{green!9}{\strut [CLS]}\colorbox{green!6}{\strut the}\colorbox{red!14}{\strut music}\colorbox{red!0}{\strut was}\colorbox{green!60}{\strut too}\colorbox{red!1}{\strut loud}\colorbox{red!5}{\strut ,}\colorbox{red!13}{\strut and}\colorbox{green!15}{\strut the}\colorbox{red!69}{\strut decorations}\colorbox{red!18}{\strut were}\colorbox{red!24}{\strut taste}\colorbox{green!37}{\strut \#\#less}\colorbox{green!1}{\strut ,}\colorbox{red!5}{\strut but}\colorbox{green!13}{\strut they}\colorbox{green!5}{\strut had}\colorbox{green!67}{\strut friendly}\colorbox{green!11}{\strut waiter}\colorbox{green!21}{\strut \#\#s}\colorbox{green!7}{\strut and}\colorbox{red!100}{\strut delicious}\colorbox{red!14}{\strut pasta}\colorbox{green!9}{\strut [SEP]} \\
& & food & $-0.88$ & \colorbox{red!5}{\strut [CLS]}\colorbox{red!3}{\strut the}\colorbox{red!1}{\strut music}\colorbox{red!3}{\strut was}\colorbox{red!3}{\strut too}\colorbox{red!2}{\strut loud}\colorbox{red!5}{\strut ,}\colorbox{red!3}{\strut and}\colorbox{red!9}{\strut the}\colorbox{green!9}{\strut decorations}\colorbox{green!1}{\strut were}\colorbox{red!4}{\strut taste}\colorbox{red!4}{\strut \#\#less}\colorbox{red!4}{\strut ,}\colorbox{green!2}{\strut but}\colorbox{red!8}{\strut they}\colorbox{red!6}{\strut had}\colorbox{red!13}{\strut friendly}\colorbox{red!26}{\strut waiter}\colorbox{red!11}{\strut \#\#s}\colorbox{red!2}{\strut and}\colorbox{green!100}{\strut delicious}\colorbox{green!13}{\strut pasta}\colorbox{red!4}{\strut [SEP]} \\
& & noise & $-1.34$ & \colorbox{green!27}{\strut [CLS]}\colorbox{green!18}{\strut the}\colorbox{green!5}{\strut music}\colorbox{red!1}{\strut was}\colorbox{red!58}{\strut too}\colorbox{red!100}{\strut loud}\colorbox{green!7}{\strut ,}\colorbox{green!16}{\strut and}\colorbox{green!20}{\strut the}\colorbox{green!27}{\strut decorations}\colorbox{green!9}{\strut were}\colorbox{green!42}{\strut taste}\colorbox{green!3}{\strut \#\#less}\colorbox{green!6}{\strut ,}\colorbox{red!3}{\strut but}\colorbox{green!6}{\strut they}\colorbox{green!13}{\strut had}\colorbox{red!38}{\strut friendly}\colorbox{red!60}{\strut waiter}\colorbox{green!9}{\strut \#\#s}\colorbox{green!8}{\strut and}\colorbox{green!15}{\strut delicious}\colorbox{green!7}{\strut pasta}\colorbox{green!18}{\strut [SEP]} \\
& & service & $+1.75$ & \colorbox{red!9}{\strut [CLS]}\colorbox{red!6}{\strut the}\colorbox{green!7}{\strut music}\colorbox{red!2}{\strut was}\colorbox{green!8}{\strut too}\colorbox{green!3}{\strut loud}\colorbox{red!4}{\strut ,}\colorbox{red!13}{\strut and}\colorbox{red!12}{\strut the}\colorbox{red!3}{\strut decorations}\colorbox{red!14}{\strut were}\colorbox{red!8}{\strut taste}\colorbox{red!11}{\strut \#\#less}\colorbox{red!5}{\strut ,}\colorbox{green!3}{\strut but}\colorbox{red!4}{\strut they}\colorbox{red!9}{\strut had}\colorbox{green!99}{\strut friendly}\colorbox{green!43}{\strut waiter}\colorbox{green!4}{\strut \#\#s}\colorbox{red!7}{\strut and}\colorbox{red!34}{\strut delicious}\colorbox{red!13}{\strut pasta}\colorbox{red!11}{\strut [SEP]} \\
\bottomrule
\end{tabular}}
  \caption{Additional visualizations of word importance scores using Integrated Gradient (IG) by restricting gradients flow through corresponding intervention site of the targeted concept. This table extends \tabref{tab:ig} in the main text.}
  \label{tab:ig-full}
\end{table*}

\Tabref{tab:ig-full} extends \tabref{tab:ig} in our main text with additional ablation studies on our training objectives.

\subsection{Model Debiasing}\label{app:debias}

Being able to accurately predict outputs for counterfactual inputs enables explanation methods to faithfully debias a model with regard to a desired concept. For instance, with CEBaB, debiasing a concept (e.g., ``food'') is equivalent to estimating the counterfactual output when we set the concept label for a concept to be \emph{unknown}. 

In this section, we briefly study the extent to which the \CPMHI{} can function as a debiasing method. To debias a concept, we enforce the sampled source input $\VarSourceInput$ as in \Eqnref{eq:cpm-hi:counterfactual} to have \emph{unknown} as its concept label for the concept to be debiased.

To show our methods can faithfully debias a targeted concept, we evaluate the correlations between the predicted overall sentiment label for sentences and the concept labels for each concept. Without any debiasing technique, we expect concept labels to be highly correlated with the overall sentiment label (e.g., if \emph{food} is positive, it is more likely that the overall sentiment is positive). We use \CPMHI{} trained for the \texttt{BERT} model architecture as an example, and use examples in the test set.

\begin{figure*}
     \centering
     \begin{subfigure}[b]{\textwidth}
         \centering
         \includegraphics[width=\textwidth]{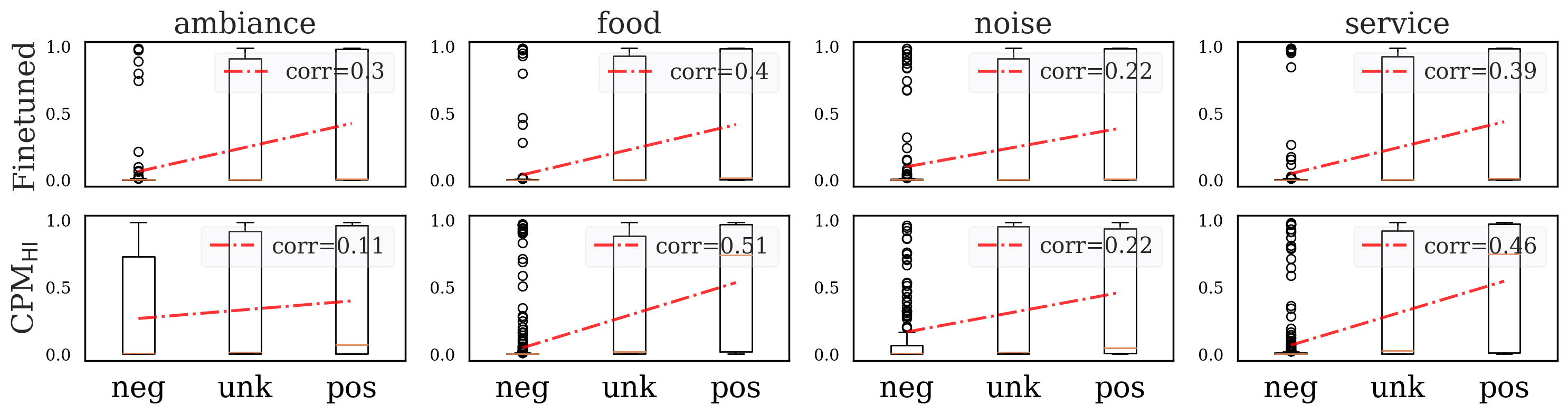}
         \caption{Visualization for debiasing the \emph{ambiance} concept.}
         \label{fig:gradients-debiasing-ambiance}
     \end{subfigure}

     \begin{subfigure}[b]{\textwidth}
         \centering
         \includegraphics[width=\textwidth]{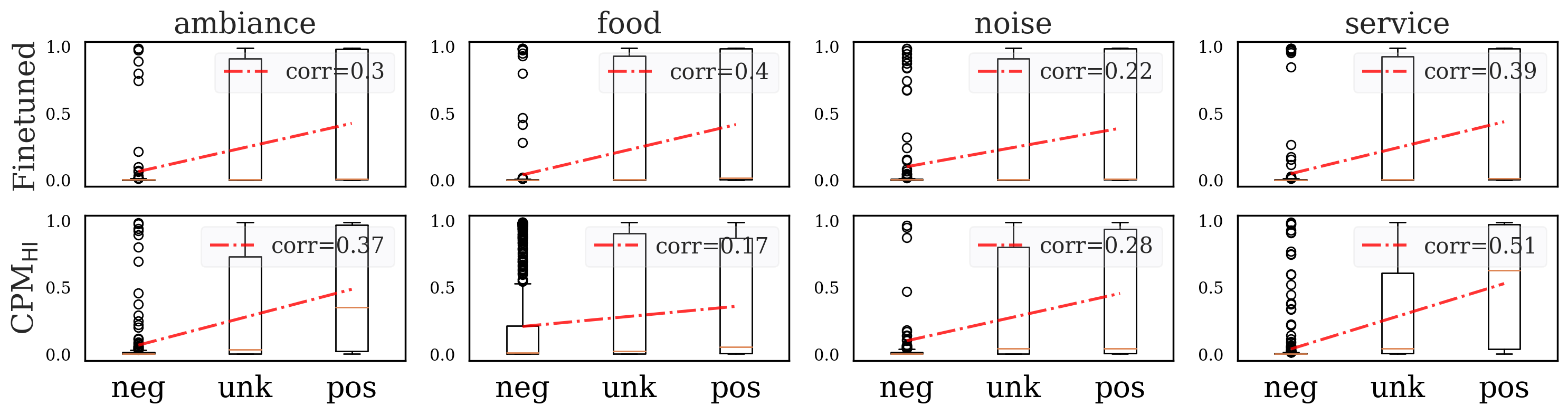}
         \caption{Visualization for debiasing the \emph{food} concept.}
         \label{fig:gradients-debiasing-food}
     \end{subfigure}

     \begin{subfigure}[b]{\textwidth}
         \centering
         \includegraphics[width=\textwidth]{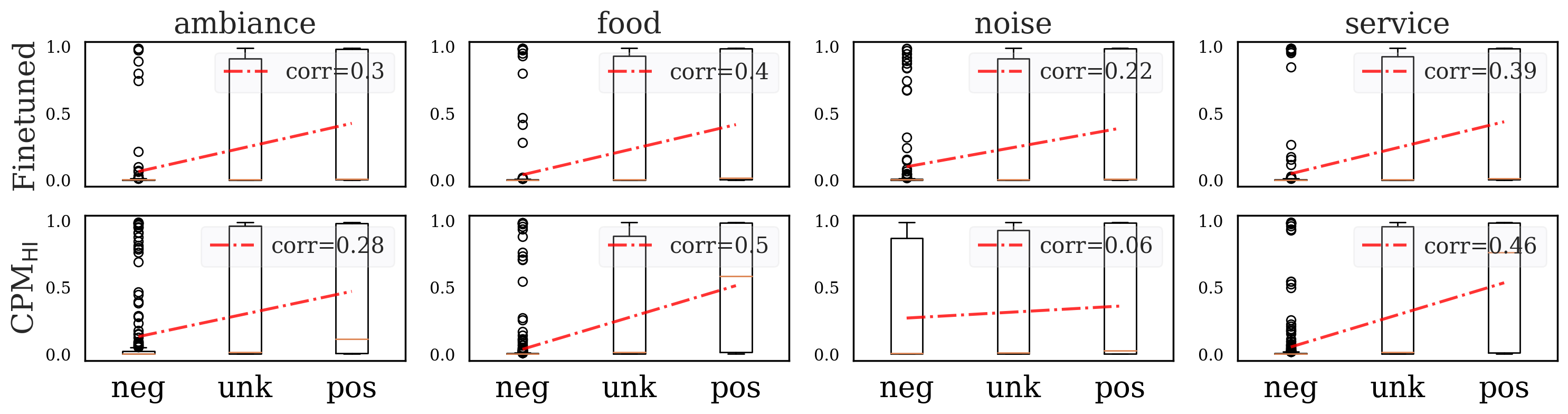}
         \caption{Visualization for debiasing the \emph{noise} concept.}
         \label{fig:gradients-debiasing-noise}
     \end{subfigure}

     \begin{subfigure}[b]{\textwidth}
         \centering
         \includegraphics[width=\textwidth]{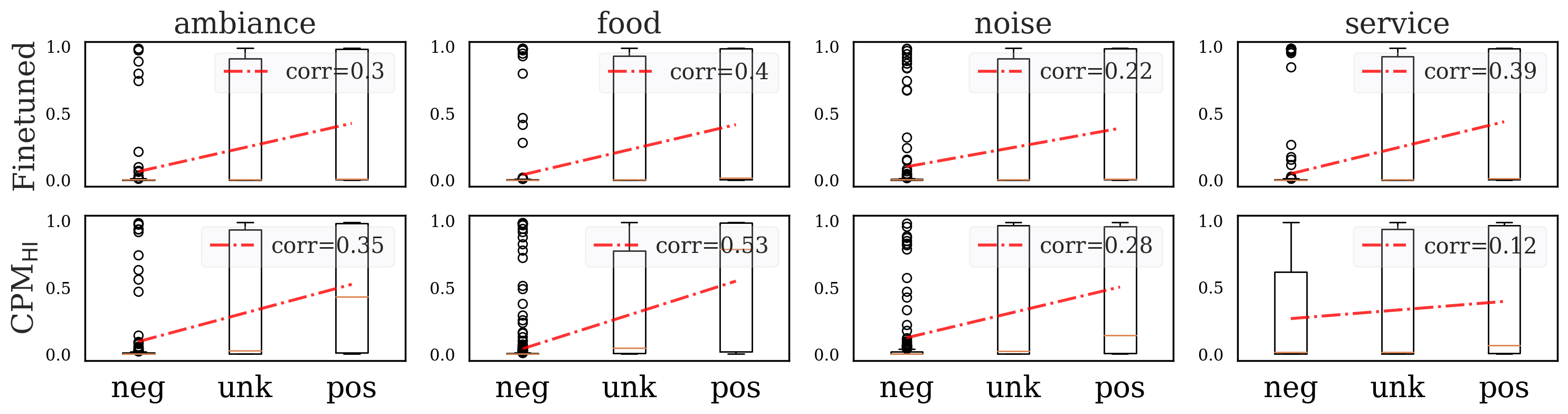}
         \caption{Visualization for debiasing the \emph{service} concept.}
         \label{fig:gradients-debiasing-service}
     \end{subfigure}
        \caption{Debiasing visualizations for different concepts of a $\text{\OurMethodAbbr{}}_{\text{\NonInputLevelAbbr{}}}$ with \texttt{BERT} model architecture. Individual plots are correlation plots between concept labels of a concept and the overall sentence sentiment label.}
        \label{fig:gradients-debiasing-all}
\end{figure*}

\Figref{fig:gradients-debiasing-all} shows correlation plots for the original \texttt{Finetuned} model as well as \CPMHI{}. As expected, the correlation of the \emph{food} concept is weakened through the debiasing pipeline by 57.50\%. Our results also suggest that correlations of other concepts are affected, which suggests a future research direction focused on minimizing the impact of the debiasing pipeline on irrelevant concepts. We include results for the remaining concepts in the \appref{app:debias}.

\Figref{fig:gradients-debiasing-ambiance} to \figref{fig:gradients-debiasing-service} show debiasing visualizations for three concepts: \emph{ambiance}, \emph{noise} and \emph{service}. We use a \CPMHI{} for the \texttt{BERT} model architecture as an example. We calculate the distributions with examples in the test set. 

\subsection{Learning Dynamics}\label{app:training-loss}

\begin{figure*}[!ht]
    \centering
    \includegraphics[width=\textwidth]{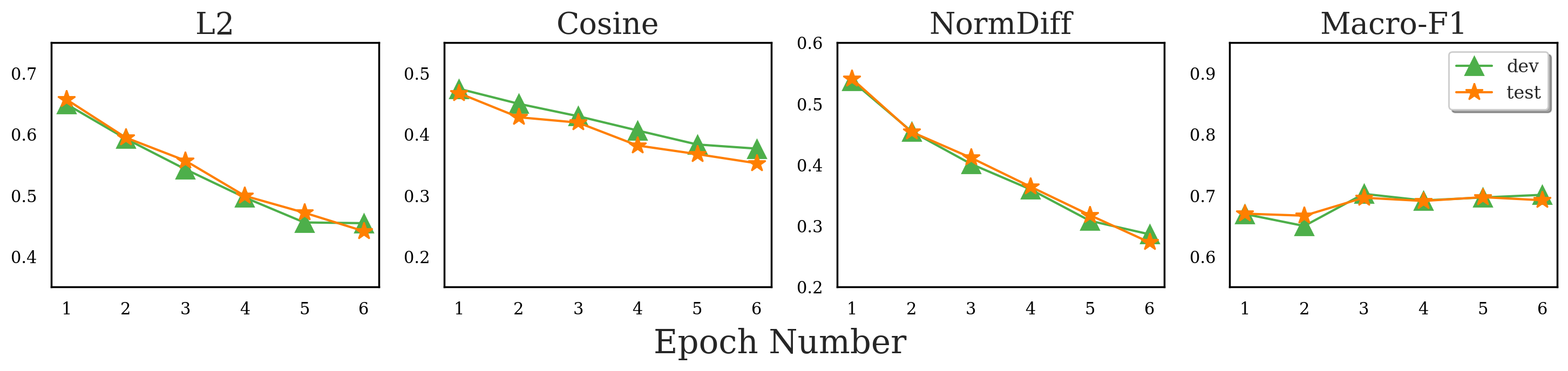}
    \caption{CEBaB scores measured in three different metrics on the dev and the test sets for a $\CPMHI{}$ with the \texttt{BERT} architectures for different training epochs. Task performance as \texttt{Macro-F1} score is reported.}
    \label{fig:metrics-epoch}
\end{figure*}

\Figref{fig:metrics-epoch} shows three different metrics measured on the dev and the test sets for a \CPMHI{} trained for the \texttt{BERT} model architecture as an example. Since we use $\text{COS}_{\texttt{ICaCE}}$ on the dev set to early stop our training process, we find our \CPMHI{} reaches a local minimum on $\text{COS}_{\texttt{ICaCE}}$ while $\text{L2}_{\texttt{ICaCE}}$ and $\text{NormDiff}_{\texttt{ICaCE}}$ are still trending downward. This suggests future research may need to choose desired metrics to optimize for during training, for early stopping to reach the best performing model.

\begin{table*}[ht]
\centering
\resizebox{0.99\linewidth}{!}{%
  \centering
  \setlength{\tabcolsep}{12pt}
  \begin{tabular}{lllcc}
\toprule
Epoch & Predicted & Concept & Score & Word Importance \\
\midrule
\multirow{5}{*}{1} & \multirow{5}{*}{\emph{neutral}}
& ambiance & $-0.17$ & \colorbox{green!5}{\strut [CLS]}\colorbox{red!3}{\strut the}\colorbox{green!1}{\strut music}\colorbox{red!27}{\strut was}\colorbox{red!53}{\strut too}\colorbox{red!11}{\strut loud}\colorbox{green!5}{\strut ,}\colorbox{red!2}{\strut and}\colorbox{green!6}{\strut the}\colorbox{green!16}{\strut decorations}\colorbox{red!13}{\strut were}\colorbox{green!11}{\strut taste}\colorbox{red!13}{\strut \#\#less}\colorbox{red!3}{\strut ,}\colorbox{green!3}{\strut but}\colorbox{red!5}{\strut they}\colorbox{red!25}{\strut had}\colorbox{green!27}{\strut friendly}\colorbox{red!24}{\strut waiter}\colorbox{red!19}{\strut \#\#s}\colorbox{red!2}{\strut and}\colorbox{green!100}{\strut delicious}\colorbox{green!19}{\strut pasta}\colorbox{green!5}{\strut [SEP]} \\
& & food & $+0.66$ & \colorbox{red!3}{\strut [CLS]}\colorbox{red!4}{\strut the}\colorbox{green!0}{\strut music}\colorbox{red!7}{\strut was}\colorbox{red!2}{\strut too}\colorbox{green!4}{\strut loud}\colorbox{red!10}{\strut ,}\colorbox{red!5}{\strut and}\colorbox{red!4}{\strut the}\colorbox{red!2}{\strut decorations}\colorbox{red!3}{\strut were}\colorbox{red!2}{\strut taste}\colorbox{green!2}{\strut \#\#less}\colorbox{red!6}{\strut ,}\colorbox{green!8}{\strut but}\colorbox{red!3}{\strut they}\colorbox{red!10}{\strut had}\colorbox{green!1}{\strut friendly}\colorbox{red!19}{\strut waiter}\colorbox{red!9}{\strut \#\#s}\colorbox{red!2}{\strut and}\colorbox{green!99}{\strut delicious}\colorbox{red!16}{\strut pasta}\colorbox{red!4}{\strut [SEP]} \\
& & noise & $-0.32$ & \colorbox{green!18}{\strut [CLS]}\colorbox{green!17}{\strut the}\colorbox{red!12}{\strut music}\colorbox{green!14}{\strut was}\colorbox{red!50}{\strut too}\colorbox{red!42}{\strut loud}\colorbox{green!3}{\strut ,}\colorbox{green!22}{\strut and}\colorbox{green!12}{\strut the}\colorbox{green!12}{\strut decorations}\colorbox{green!16}{\strut were}\colorbox{green!6}{\strut taste}\colorbox{red!21}{\strut \#\#less}\colorbox{red!2}{\strut ,}\colorbox{red!3}{\strut but}\colorbox{green!25}{\strut they}\colorbox{green!35}{\strut had}\colorbox{green!46}{\strut friendly}\colorbox{green!5}{\strut waiter}\colorbox{red!0}{\strut \#\#s}\colorbox{red!21}{\strut and}\colorbox{red!100}{\strut delicious}\colorbox{green!6}{\strut pasta}\colorbox{green!12}{\strut [SEP]} \\
& & service & $+0.05$ & \colorbox{red!0}{\strut [CLS]}\colorbox{green!3}{\strut the}\colorbox{green!1}{\strut music}\colorbox{red!19}{\strut was}\colorbox{red!20}{\strut too}\colorbox{red!0}{\strut loud}\colorbox{red!13}{\strut ,}\colorbox{red!2}{\strut and}\colorbox{red!1}{\strut the}\colorbox{green!2}{\strut decorations}\colorbox{red!14}{\strut were}\colorbox{red!5}{\strut taste}\colorbox{red!16}{\strut \#\#less}\colorbox{red!9}{\strut ,}\colorbox{green!8}{\strut but}\colorbox{green!0}{\strut they}\colorbox{red!18}{\strut had}\colorbox{green!29}{\strut friendly}\colorbox{red!16}{\strut waiter}\colorbox{red!14}{\strut \#\#s}\colorbox{green!2}{\strut and}\colorbox{green!99}{\strut delicious}\colorbox{green!5}{\strut pasta}\colorbox{green!1}{\strut [SEP]} \\
\midrule
\multirow{5}{*}{2} & \multirow{5}{*}{\emph{neutral}}
& ambiance & $-0.25$ & \colorbox{green!3}{\strut [CLS]}\colorbox{green!6}{\strut the}\colorbox{green!2}{\strut music}\colorbox{red!9}{\strut was}\colorbox{red!13}{\strut too}\colorbox{red!14}{\strut loud}\colorbox{green!3}{\strut ,}\colorbox{green!2}{\strut and}\colorbox{green!1}{\strut the}\colorbox{green!17}{\strut decorations}\colorbox{red!18}{\strut were}\colorbox{green!4}{\strut taste}\colorbox{red!4}{\strut \#\#less}\colorbox{red!5}{\strut ,}\colorbox{red!0}{\strut but}\colorbox{green!4}{\strut they}\colorbox{red!5}{\strut had}\colorbox{green!11}{\strut friendly}\colorbox{green!23}{\strut waiter}\colorbox{green!3}{\strut \#\#s}\colorbox{red!8}{\strut and}\colorbox{red!100}{\strut delicious}\colorbox{green!86}{\strut pasta}\colorbox{green!6}{\strut [SEP]} \\
& & food & $+1.54$ & \colorbox{red!0}{\strut [CLS]}\colorbox{red!4}{\strut the}\colorbox{red!8}{\strut music}\colorbox{red!5}{\strut was}\colorbox{green!13}{\strut too}\colorbox{green!10}{\strut loud}\colorbox{red!11}{\strut ,}\colorbox{red!7}{\strut and}\colorbox{red!4}{\strut the}\colorbox{green!4}{\strut decorations}\colorbox{green!8}{\strut were}\colorbox{red!11}{\strut taste}\colorbox{green!4}{\strut \#\#less}\colorbox{red!0}{\strut ,}\colorbox{green!0}{\strut but}\colorbox{red!4}{\strut they}\colorbox{red!13}{\strut had}\colorbox{red!9}{\strut friendly}\colorbox{red!34}{\strut waiter}\colorbox{red!9}{\strut \#\#s}\colorbox{green!6}{\strut and}\colorbox{green!100}{\strut delicious}\colorbox{red!20}{\strut pasta}\colorbox{red!3}{\strut [SEP]} \\
& & noise & $-0.24$ & \colorbox{green!17}{\strut [CLS]}\colorbox{green!12}{\strut the}\colorbox{red!24}{\strut music}\colorbox{green!58}{\strut was}\colorbox{red!17}{\strut too}\colorbox{red!43}{\strut loud}\colorbox{green!8}{\strut ,}\colorbox{red!2}{\strut and}\colorbox{red!1}{\strut the}\colorbox{red!2}{\strut decorations}\colorbox{green!29}{\strut were}\colorbox{green!1}{\strut taste}\colorbox{red!5}{\strut \#\#less}\colorbox{green!6}{\strut ,}\colorbox{red!5}{\strut but}\colorbox{green!37}{\strut they}\colorbox{green!15}{\strut had}\colorbox{green!36}{\strut friendly}\colorbox{red!2}{\strut waiter}\colorbox{red!6}{\strut \#\#s}\colorbox{red!4}{\strut and}\colorbox{red!100}{\strut delicious}\colorbox{red!18}{\strut pasta}\colorbox{green!11}{\strut [SEP]} \\
& & service & $+0.02$ & \colorbox{green!4}{\strut [CLS]}\colorbox{green!1}{\strut the}\colorbox{green!16}{\strut music}\colorbox{red!18}{\strut was}\colorbox{red!2}{\strut too}\colorbox{red!6}{\strut loud}\colorbox{red!10}{\strut ,}\colorbox{red!5}{\strut and}\colorbox{green!3}{\strut the}\colorbox{red!1}{\strut decorations}\colorbox{red!19}{\strut were}\colorbox{red!0}{\strut taste}\colorbox{red!8}{\strut \#\#less}\colorbox{red!15}{\strut ,}\colorbox{green!4}{\strut but}\colorbox{green!8}{\strut they}\colorbox{red!9}{\strut had}\colorbox{green!64}{\strut friendly}\colorbox{green!22}{\strut waiter}\colorbox{red!6}{\strut \#\#s}\colorbox{red!13}{\strut and}\colorbox{red!100}{\strut delicious}\colorbox{green!91}{\strut pasta}\colorbox{red!0}{\strut [SEP]} \\
\midrule
\multirow{5}{*}{3} & \multirow{5}{*}{\emph{neutral}}
& ambiance & $-0.49$ & \colorbox{green!13}{\strut [CLS]}\colorbox{green!11}{\strut the}\colorbox{green!62}{\strut music}\colorbox{red!4}{\strut was}\colorbox{red!45}{\strut too}\colorbox{red!34}{\strut loud}\colorbox{red!13}{\strut ,}\colorbox{green!1}{\strut and}\colorbox{green!15}{\strut the}\colorbox{red!5}{\strut decorations}\colorbox{red!2}{\strut were}\colorbox{red!6}{\strut taste}\colorbox{red!0}{\strut \#\#less}\colorbox{red!14}{\strut ,}\colorbox{green!21}{\strut but}\colorbox{green!15}{\strut they}\colorbox{green!10}{\strut had}\colorbox{green!65}{\strut friendly}\colorbox{green!12}{\strut waiter}\colorbox{red!1}{\strut \#\#s}\colorbox{red!12}{\strut and}\colorbox{red!100}{\strut delicious}\colorbox{red!2}{\strut pasta}\colorbox{green!13}{\strut [SEP]} \\
& & food & $+1.52$ & \colorbox{red!7}{\strut [CLS]}\colorbox{red!5}{\strut the}\colorbox{red!1}{\strut music}\colorbox{red!2}{\strut was}\colorbox{green!5}{\strut too}\colorbox{green!21}{\strut loud}\colorbox{red!3}{\strut ,}\colorbox{red!6}{\strut and}\colorbox{red!6}{\strut the}\colorbox{green!1}{\strut decorations}\colorbox{green!1}{\strut were}\colorbox{red!16}{\strut taste}\colorbox{red!15}{\strut \#\#less}\colorbox{red!2}{\strut ,}\colorbox{green!3}{\strut but}\colorbox{red!5}{\strut they}\colorbox{red!9}{\strut had}\colorbox{red!12}{\strut friendly}\colorbox{red!27}{\strut waiter}\colorbox{red!15}{\strut \#\#s}\colorbox{green!6}{\strut and}\colorbox{green!100}{\strut delicious}\colorbox{green!5}{\strut pasta}\colorbox{red!8}{\strut [SEP]} \\
& & noise & $-0.97$ & \colorbox{green!18}{\strut [CLS]}\colorbox{green!26}{\strut the}\colorbox{green!4}{\strut music}\colorbox{green!14}{\strut was}\colorbox{red!43}{\strut too}\colorbox{red!100}{\strut loud}\colorbox{green!0}{\strut ,}\colorbox{green!16}{\strut and}\colorbox{green!21}{\strut the}\colorbox{green!6}{\strut decorations}\colorbox{green!9}{\strut were}\colorbox{green!14}{\strut taste}\colorbox{green!6}{\strut \#\#less}\colorbox{red!1}{\strut ,}\colorbox{green!13}{\strut but}\colorbox{green!9}{\strut they}\colorbox{green!9}{\strut had}\colorbox{green!34}{\strut friendly}\colorbox{red!8}{\strut waiter}\colorbox{green!10}{\strut \#\#s}\colorbox{red!12}{\strut and}\colorbox{red!51}{\strut delicious}\colorbox{red!16}{\strut pasta}\colorbox{green!14}{\strut [SEP]} \\
& & service & $+0.49$ & \colorbox{red!6}{\strut [CLS]}\colorbox{green!1}{\strut the}\colorbox{green!36}{\strut music}\colorbox{red!20}{\strut was}\colorbox{red!3}{\strut too}\colorbox{green!56}{\strut loud}\colorbox{red!25}{\strut ,}\colorbox{red!11}{\strut and}\colorbox{green!1}{\strut the}\colorbox{red!13}{\strut decorations}\colorbox{red!18}{\strut were}\colorbox{red!7}{\strut taste}\colorbox{red!7}{\strut \#\#less}\colorbox{red!13}{\strut ,}\colorbox{green!12}{\strut but}\colorbox{green!1}{\strut they}\colorbox{red!14}{\strut had}\colorbox{green!100}{\strut friendly}\colorbox{green!62}{\strut waiter}\colorbox{red!12}{\strut \#\#s}\colorbox{red!23}{\strut and}\colorbox{red!63}{\strut delicious}\colorbox{red!24}{\strut pasta}\colorbox{red!6}{\strut [SEP]} \\
\midrule
\multirow{5}{*}{4} & \multirow{5}{*}{\emph{neutral}}
& ambiance & $-0.69$ & \colorbox{green!26}{\strut [CLS]}\colorbox{green!41}{\strut the}\colorbox{red!2}{\strut music}\colorbox{red!13}{\strut was}\colorbox{red!46}{\strut too}\colorbox{red!86}{\strut loud}\colorbox{red!13}{\strut ,}\colorbox{green!10}{\strut and}\colorbox{green!26}{\strut the}\colorbox{red!24}{\strut decorations}\colorbox{red!3}{\strut were}\colorbox{red!33}{\strut taste}\colorbox{green!9}{\strut \#\#less}\colorbox{red!13}{\strut ,}\colorbox{green!14}{\strut but}\colorbox{green!24}{\strut they}\colorbox{green!26}{\strut had}\colorbox{green!72}{\strut friendly}\colorbox{red!100}{\strut waiter}\colorbox{red!0}{\strut \#\#s}\colorbox{green!11}{\strut and}\colorbox{green!12}{\strut delicious}\colorbox{green!34}{\strut pasta}\colorbox{green!26}{\strut [SEP]} \\
& & food & $+1.41$ & \colorbox{red!8}{\strut [CLS]}\colorbox{red!2}{\strut the}\colorbox{red!14}{\strut music}\colorbox{red!11}{\strut was}\colorbox{green!3}{\strut too}\colorbox{green!38}{\strut loud}\colorbox{red!5}{\strut ,}\colorbox{red!11}{\strut and}\colorbox{red!5}{\strut the}\colorbox{red!2}{\strut decorations}\colorbox{green!5}{\strut were}\colorbox{red!31}{\strut taste}\colorbox{red!14}{\strut \#\#less}\colorbox{red!7}{\strut ,}\colorbox{green!12}{\strut but}\colorbox{green!4}{\strut they}\colorbox{red!9}{\strut had}\colorbox{red!7}{\strut friendly}\colorbox{red!15}{\strut waiter}\colorbox{red!7}{\strut \#\#s}\colorbox{green!3}{\strut and}\colorbox{green!100}{\strut delicious}\colorbox{red!3}{\strut pasta}\colorbox{red!8}{\strut [SEP]} \\
& & noise & $-1.92$ & \colorbox{green!10}{\strut [CLS]}\colorbox{green!9}{\strut the}\colorbox{red!17}{\strut music}\colorbox{green!10}{\strut was}\colorbox{red!24}{\strut too}\colorbox{red!100}{\strut loud}\colorbox{green!3}{\strut ,}\colorbox{green!14}{\strut and}\colorbox{green!6}{\strut the}\colorbox{green!9}{\strut decorations}\colorbox{green!9}{\strut were}\colorbox{green!13}{\strut taste}\colorbox{green!5}{\strut \#\#less}\colorbox{green!3}{\strut ,}\colorbox{green!6}{\strut but}\colorbox{green!1}{\strut they}\colorbox{green!1}{\strut had}\colorbox{green!15}{\strut friendly}\colorbox{red!2}{\strut waiter}\colorbox{green!4}{\strut \#\#s}\colorbox{red!1}{\strut and}\colorbox{green!7}{\strut delicious}\colorbox{green!2}{\strut pasta}\colorbox{green!10}{\strut [SEP]} \\
& & service & $+1.14$ & \colorbox{red!12}{\strut [CLS]}\colorbox{green!1}{\strut the}\colorbox{green!47}{\strut music}\colorbox{red!10}{\strut was}\colorbox{red!31}{\strut too}\colorbox{green!55}{\strut loud}\colorbox{red!24}{\strut ,}\colorbox{red!20}{\strut and}\colorbox{red!3}{\strut the}\colorbox{red!4}{\strut decorations}\colorbox{red!22}{\strut were}\colorbox{red!10}{\strut taste}\colorbox{red!27}{\strut \#\#less}\colorbox{red!20}{\strut ,}\colorbox{green!13}{\strut but}\colorbox{red!5}{\strut they}\colorbox{red!13}{\strut had}\colorbox{green!100}{\strut friendly}\colorbox{green!50}{\strut waiter}\colorbox{red!6}{\strut \#\#s}\colorbox{red!11}{\strut and}\colorbox{green!2}{\strut delicious}\colorbox{red!31}{\strut pasta}\colorbox{red!13}{\strut [SEP]} \\
\midrule
\multirow{5}{*}{5} & \multirow{5}{*}{\emph{neutral}}
& ambiance & $-0.77$ & \colorbox{green!14}{\strut [CLS]}\colorbox{green!24}{\strut the}\colorbox{red!5}{\strut music}\colorbox{green!4}{\strut was}\colorbox{green!36}{\strut too}\colorbox{red!10}{\strut loud}\colorbox{red!9}{\strut ,}\colorbox{green!0}{\strut and}\colorbox{green!12}{\strut the}\colorbox{red!41}{\strut decorations}\colorbox{red!1}{\strut were}\colorbox{green!46}{\strut taste}\colorbox{green!34}{\strut \#\#less}\colorbox{red!7}{\strut ,}\colorbox{red!6}{\strut but}\colorbox{green!9}{\strut they}\colorbox{green!3}{\strut had}\colorbox{green!21}{\strut friendly}\colorbox{red!76}{\strut waiter}\colorbox{red!1}{\strut \#\#s}\colorbox{red!10}{\strut and}\colorbox{red!100}{\strut delicious}\colorbox{green!45}{\strut pasta}\colorbox{green!14}{\strut [SEP]} \\
& & food & $+1.25$ & \colorbox{red!2}{\strut [CLS]}\colorbox{red!4}{\strut the}\colorbox{red!4}{\strut music}\colorbox{red!3}{\strut was}\colorbox{red!1}{\strut too}\colorbox{red!2}{\strut loud}\colorbox{red!2}{\strut ,}\colorbox{red!3}{\strut and}\colorbox{red!2}{\strut the}\colorbox{green!7}{\strut decorations}\colorbox{green!4}{\strut were}\colorbox{red!23}{\strut taste}\colorbox{red!13}{\strut \#\#less}\colorbox{red!1}{\strut ,}\colorbox{green!8}{\strut but}\colorbox{green!1}{\strut they}\colorbox{red!1}{\strut had}\colorbox{red!8}{\strut friendly}\colorbox{red!11}{\strut waiter}\colorbox{red!6}{\strut \#\#s}\colorbox{green!2}{\strut and}\colorbox{green!100}{\strut delicious}\colorbox{red!26}{\strut pasta}\colorbox{red!3}{\strut [SEP]} \\
& & noise & $-1.63$ & \colorbox{green!9}{\strut [CLS]}\colorbox{green!13}{\strut the}\colorbox{red!19}{\strut music}\colorbox{green!8}{\strut was}\colorbox{red!29}{\strut too}\colorbox{red!100}{\strut loud}\colorbox{green!6}{\strut ,}\colorbox{green!12}{\strut and}\colorbox{green!7}{\strut the}\colorbox{green!1}{\strut decorations}\colorbox{green!7}{\strut were}\colorbox{green!12}{\strut taste}\colorbox{green!5}{\strut \#\#less}\colorbox{green!3}{\strut ,}\colorbox{green!7}{\strut but}\colorbox{green!2}{\strut they}\colorbox{green!0}{\strut had}\colorbox{green!18}{\strut friendly}\colorbox{green!9}{\strut waiter}\colorbox{green!7}{\strut \#\#s}\colorbox{green!2}{\strut and}\colorbox{green!1}{\strut delicious}\colorbox{green!0}{\strut pasta}\colorbox{green!9}{\strut [SEP]} \\
& & service & $+1.28$ & \colorbox{red!9}{\strut [CLS]}\colorbox{red!6}{\strut the}\colorbox{green!10}{\strut music}\colorbox{red!14}{\strut was}\colorbox{red!8}{\strut too}\colorbox{green!24}{\strut loud}\colorbox{red!14}{\strut ,}\colorbox{red!10}{\strut and}\colorbox{red!4}{\strut the}\colorbox{red!13}{\strut decorations}\colorbox{red!16}{\strut were}\colorbox{green!0}{\strut taste}\colorbox{red!7}{\strut \#\#less}\colorbox{red!13}{\strut ,}\colorbox{green!11}{\strut but}\colorbox{red!14}{\strut they}\colorbox{red!17}{\strut had}\colorbox{green!100}{\strut friendly}\colorbox{green!38}{\strut waiter}\colorbox{red!3}{\strut \#\#s}\colorbox{red!7}{\strut and}\colorbox{red!11}{\strut delicious}\colorbox{red!2}{\strut pasta}\colorbox{red!8}{\strut [SEP]} \\
\midrule
\multirow{5}{*}{6} & \multirow{5}{*}{\emph{neutral}}
& ambiance & $-0.66$ & \colorbox{green!31}{\strut [CLS]}\colorbox{green!31}{\strut the}\colorbox{green!8}{\strut music}\colorbox{red!33}{\strut was}\colorbox{green!98}{\strut too}\colorbox{red!51}{\strut loud}\colorbox{red!42}{\strut ,}\colorbox{red!1}{\strut and}\colorbox{green!23}{\strut the}\colorbox{red!100}{\strut decorations}\colorbox{red!27}{\strut were}\colorbox{red!6}{\strut taste}\colorbox{green!36}{\strut \#\#less}\colorbox{red!46}{\strut ,}\colorbox{red!13}{\strut but}\colorbox{green!31}{\strut they}\colorbox{red!4}{\strut had}\colorbox{green!91}{\strut friendly}\colorbox{red!62}{\strut waiter}\colorbox{green!8}{\strut \#\#s}\colorbox{green!25}{\strut and}\colorbox{red!73}{\strut delicious}\colorbox{green!46}{\strut pasta}\colorbox{green!31}{\strut [SEP]} \\
& & food & $+0.62$ & \colorbox{red!1}{\strut [CLS]}\colorbox{red!2}{\strut the}\colorbox{red!5}{\strut music}\colorbox{red!6}{\strut was}\colorbox{green!9}{\strut too}\colorbox{red!7}{\strut loud}\colorbox{red!4}{\strut ,}\colorbox{red!2}{\strut and}\colorbox{red!8}{\strut the}\colorbox{green!13}{\strut decorations}\colorbox{green!11}{\strut were}\colorbox{red!15}{\strut taste}\colorbox{red!11}{\strut \#\#less}\colorbox{red!4}{\strut ,}\colorbox{green!4}{\strut but}\colorbox{red!2}{\strut they}\colorbox{red!0}{\strut had}\colorbox{red!18}{\strut friendly}\colorbox{red!32}{\strut waiter}\colorbox{red!13}{\strut \#\#s}\colorbox{green!5}{\strut and}\colorbox{green!100}{\strut delicious}\colorbox{red!5}{\strut pasta}\colorbox{red!1}{\strut [SEP]} \\
& & noise & $-0.90$ & \colorbox{green!13}{\strut [CLS]}\colorbox{green!17}{\strut the}\colorbox{red!20}{\strut music}\colorbox{green!5}{\strut was}\colorbox{red!53}{\strut too}\colorbox{red!100}{\strut loud}\colorbox{red!7}{\strut ,}\colorbox{green!18}{\strut and}\colorbox{green!8}{\strut the}\colorbox{red!6}{\strut decorations}\colorbox{green!16}{\strut were}\colorbox{green!3}{\strut taste}\colorbox{green!1}{\strut \#\#less}\colorbox{red!4}{\strut ,}\colorbox{green!7}{\strut but}\colorbox{green!3}{\strut they}\colorbox{green!4}{\strut had}\colorbox{green!30}{\strut friendly}\colorbox{green!41}{\strut waiter}\colorbox{green!10}{\strut \#\#s}\colorbox{red!6}{\strut and}\colorbox{green!17}{\strut delicious}\colorbox{red!17}{\strut pasta}\colorbox{green!14}{\strut [SEP]} \\
& & service & $+2.14$ & \colorbox{red!17}{\strut [CLS]}\colorbox{red!12}{\strut the}\colorbox{green!29}{\strut music}\colorbox{red!15}{\strut was}\colorbox{red!24}{\strut too}\colorbox{green!29}{\strut loud}\colorbox{red!18}{\strut ,}\colorbox{red!21}{\strut and}\colorbox{red!6}{\strut the}\colorbox{red!5}{\strut decorations}\colorbox{red!18}{\strut were}\colorbox{red!13}{\strut taste}\colorbox{red!11}{\strut \#\#less}\colorbox{red!17}{\strut ,}\colorbox{green!14}{\strut but}\colorbox{red!9}{\strut they}\colorbox{red!16}{\strut had}\colorbox{green!100}{\strut friendly}\colorbox{green!46}{\strut waiter}\colorbox{green!5}{\strut \#\#s}\colorbox{red!14}{\strut and}\colorbox{green!30}{\strut delicious}\colorbox{red!15}{\strut pasta}\colorbox{red!17}{\strut [SEP]} \\
\midrule
\multirow{5}{*}{\textbf{\CPMHI}} & \multirow{5}{*}{\emph{neutral}}
& ambiance & $-0.61$ & \colorbox{green!9}{\strut [CLS]}\colorbox{green!6}{\strut the}\colorbox{red!14}{\strut music}\colorbox{red!0}{\strut was}\colorbox{green!60}{\strut too}\colorbox{red!1}{\strut loud}\colorbox{red!5}{\strut ,}\colorbox{red!13}{\strut and}\colorbox{green!15}{\strut the}\colorbox{red!69}{\strut decorations}\colorbox{red!18}{\strut were}\colorbox{red!24}{\strut taste}\colorbox{green!37}{\strut \#\#less}\colorbox{green!1}{\strut ,}\colorbox{red!5}{\strut but}\colorbox{green!13}{\strut they}\colorbox{green!5}{\strut had}\colorbox{green!67}{\strut friendly}\colorbox{green!11}{\strut waiter}\colorbox{green!21}{\strut \#\#s}\colorbox{green!7}{\strut and}\colorbox{red!100}{\strut delicious}\colorbox{red!14}{\strut pasta}\colorbox{green!9}{\strut [SEP]} \\
& & food & $-0.88$ & \colorbox{red!5}{\strut [CLS]}\colorbox{red!3}{\strut the}\colorbox{red!1}{\strut music}\colorbox{red!3}{\strut was}\colorbox{red!3}{\strut too}\colorbox{red!2}{\strut loud}\colorbox{red!5}{\strut ,}\colorbox{red!3}{\strut and}\colorbox{red!9}{\strut the}\colorbox{green!9}{\strut decorations}\colorbox{green!1}{\strut were}\colorbox{red!4}{\strut taste}\colorbox{red!4}{\strut \#\#less}\colorbox{red!4}{\strut ,}\colorbox{green!2}{\strut but}\colorbox{red!8}{\strut they}\colorbox{red!6}{\strut had}\colorbox{red!13}{\strut friendly}\colorbox{red!26}{\strut waiter}\colorbox{red!11}{\strut \#\#s}\colorbox{red!2}{\strut and}\colorbox{green!100}{\strut delicious}\colorbox{green!13}{\strut pasta}\colorbox{red!4}{\strut [SEP]} \\
& & noise & $-1.34$ & \colorbox{green!27}{\strut [CLS]}\colorbox{green!18}{\strut the}\colorbox{green!5}{\strut music}\colorbox{red!1}{\strut was}\colorbox{red!58}{\strut too}\colorbox{red!100}{\strut loud}\colorbox{green!7}{\strut ,}\colorbox{green!16}{\strut and}\colorbox{green!20}{\strut the}\colorbox{green!27}{\strut decorations}\colorbox{green!9}{\strut were}\colorbox{green!42}{\strut taste}\colorbox{green!3}{\strut \#\#less}\colorbox{green!6}{\strut ,}\colorbox{red!3}{\strut but}\colorbox{green!6}{\strut they}\colorbox{green!13}{\strut had}\colorbox{red!38}{\strut friendly}\colorbox{red!60}{\strut waiter}\colorbox{green!9}{\strut \#\#s}\colorbox{green!8}{\strut and}\colorbox{green!15}{\strut delicious}\colorbox{green!7}{\strut pasta}\colorbox{green!18}{\strut [SEP]} \\
& & service & $+1.75$ & \colorbox{red!9}{\strut [CLS]}\colorbox{red!6}{\strut the}\colorbox{green!7}{\strut music}\colorbox{red!2}{\strut was}\colorbox{green!8}{\strut too}\colorbox{green!3}{\strut loud}\colorbox{red!4}{\strut ,}\colorbox{red!13}{\strut and}\colorbox{red!12}{\strut the}\colorbox{red!3}{\strut decorations}\colorbox{red!14}{\strut were}\colorbox{red!8}{\strut taste}\colorbox{red!11}{\strut \#\#less}\colorbox{red!5}{\strut ,}\colorbox{green!3}{\strut but}\colorbox{red!4}{\strut they}\colorbox{red!9}{\strut had}\colorbox{green!99}{\strut friendly}\colorbox{green!43}{\strut waiter}\colorbox{green!4}{\strut \#\#s}\colorbox{red!7}{\strut and}\colorbox{red!34}{\strut delicious}\colorbox{red!13}{\strut pasta}\colorbox{red!11}{\strut [SEP]} \\
\bottomrule
\end{tabular}}
  \caption{Visualizations of word importance scores using Integrated Gradient (IG), using the same methods as in \tabref{tab:ig} and \tabref{tab:ig-full}.}
  \label{tab:ig-epoch}
\end{table*}

\Tabref{tab:ig-epoch} visualizations of word importance scores using our version of Integrated Gradient (IG). Different from \tabref{tab:ig} and \tabref{tab:ig-full}, which show the visualizations of our optimized model, we show a per-epoch result for for \CPMHI{}, followed with our best model appended at the end. Our results suggest that early checkpoints in the training process focus at drastically different input words comparing to later checkpoints, though all models predict \emph{neutral} for this given sentence. In addition, gradient aggregations over input words are rather stable towards the end the training. More importantly, \CPMHI{} learns how to highlight words that are semantically related to each concept gradually. For instance, we can see a clear trend of emphasising the word ``decorations'' for the \emph{ambiance} concept throughout the training process. This suggests that our training procedure induces causally motivated gradients over input words gradually through the training process.

\end{document}